\documentclass[11pt]{article}
\usepackage[final]{acl}
\usepackage{times}
\usepackage{latexsym}
\usepackage[T1]{fontenc}
\usepackage[utf8]{inputenc}
\usepackage{microtype}
\usepackage{inconsolata}
\usepackage{graphicx}
\usepackage{booktabs}
\usepackage{subcaption}
\usepackage{multirow}
\usepackage{booktabs}
\usepackage{xcolor}
\usepackage{amssymb}
\usepackage{hyperref}
\usepackage[most]{tcolorbox}
\usepackage{xcolor}
\usepackage{amsmath}
\usepackage{pifont}
\usepackage{xurl}
\newcommand{\cmark}{\ding{51}} 
\newcommand{\xmark}{\ding{55}} 
\usepackage{listings}
\usepackage{xcolor}

\title{CCRV-Bench: Constraint-Based Evaluation of Causal Reasoning in Vision-Language Models}

\title{CCRV-Bench: Constraint-Based Evaluation of Causal Reasoning
in Vision-Language Models}

\author{
  \textbf{Linyuan Gao\textsuperscript{1}} \quad
  \textbf{Yuan Wu\textsuperscript{1}}\thanks{Corresponding author.} \quad
  \textbf{Yi Chang\textsuperscript{1,2,3}}
  \\
  \textsuperscript{1}School of Artificial Intelligence, Jilin University
  \\
  \textsuperscript{2}Engineering Research Center of Knowledge-Driven
  \\
  Human-Machine Intelligence, Jilin University
  \\
  \textsuperscript{3}International Center of Future Science, Jilin University
  \\
  Changchun 130012, China
  \\
  \texttt{lygao25@mails.jlu.edu.cn}
  \\
  \texttt{yuanwu@jlu.edu.cn} \quad
  \texttt{yichang@jlu.edu.cn}
}

\begin{document}
\maketitle
\begin{abstract}
Vision-language models (VLMs) have demonstrated excellent performance in visual tasks, but their visual causal reasoning capabilities still lack reliable evaluation. Existing evaluations struggle to distinguish whether a model is performing causal reasoning based on visual evidence or relying on statistical correlations for shortcut learning, thereby potentially overestimating their actual capabilities. This paper proposes CCRV-Bench, a constraint-driven visual causal reasoning benchmark for single-image physical scenarios. We construct an orthogonal framework that evaluates four causal task dimensions: causal relation discovery, state prediction, causal diagnosis, and intervention-outcome prediction. We further introduce entity symbolization, spatial grounding, the factual adversarial constraint, and minimalist output constraints to reduce shortcut cues while preserving the physical commonsense required by the task. Experiments across 14 multimodal models show that constraint sensitivity is task- and model-dependent: intervention-outcome prediction has the largest average effective degradation among the four causal tasks, spatial grounding is the most damaging constraint on average, and the factual adversarial constraint improves DCR for all evaluated models. These results show that unconstrained performance does not determine constrained robustness and that a single aggregate score can obscure distinct failures in causal identification, spatial grounding, and constraint-compliant expression. CCRV-Bench provides a standardized framework for diagnosing image-grounded causal reasoning under controlled constraints. The code is available at \url{https://github.com/0815linyuan/CCRVBench}

\end{abstract}
\section{Introduction}
\begin{table*}[t]
\centering

\resizebox{\textwidth}{!}{
\begin{tabular}{lccccccccc}
\toprule

\multirow{2}{*}{\textbf{Benchmark}} 
& \multicolumn{4}{c}{\textbf{Causal Capability}} 
& \multicolumn{3}{c}{\textbf{Constraint}} 
& \textbf{Input} 
& \textbf{Process} \\

\cmidrule(lr){2-5} \cmidrule(lr){6-8}

& \textbf{Und.} 
& \textbf{Pred.} 
& \textbf{Attr.} 
& \textbf{Interv.} 
& \textbf{Ground.} 
& \textbf{Factual Adversarial Constraint} 
& \textbf{Shortcut} 
& \textbf{Modality} 
& \textbf{Proc. Eval} \\

\midrule

CELLO~\cite{chen2024cello}          & \cmark & \cmark & \xmark & \cmark & \xmark & \xmark & \xmark & Image        & \xmark \\
CausalVLBench~\cite{komanduri2025causalvlbench}   & \xmark & \cmark & \cmark & \cmark & \xmark & \xmark & \xmark & Image       & \xmark \\
CausalVQA~\cite{foss2025causalvqa}       & \cmark & \cmark & \cmark & \cmark & \xmark & \xmark & \xmark & video       & \xmark \\
GQA~\cite{hudson2019gqa}             & \xmark & \xmark & \cmark & \xmark & \xmark & \xmark & \xmark & Image       & \xmark \\
CFBench~\cite{zhang2025cfbench}         & \xmark & \xmark & \xmark & \xmark & \xmark & \xmark & \cmark & Text       & \xmark \\
CCR-Bench~\cite{xue2026ccr}       & \xmark & \xmark & \xmark & \xmark & \xmark & \xmark & \cmark & Text        & \xmark \\
CAUSAL3D~\cite{liu2025causal3d}        & \cmark & \xmark & \xmark & \cmark & \xmark & \xmark & \xmark & Multi-image & \xmark \\
ComplexBench~\cite{wen2024benchmarking}    & \xmark & \xmark & \xmark & \xmark & \xmark & \xmark & \cmark & Text       & \cmark \\
CausalBench~\cite{zhou2024causalbench}     & \cmark & \cmark & \xmark & \xmark & \xmark & \xmark & \cmark & Text        & \xmark \\

\midrule

\textbf{CCRV-Bench (Ours)} 
& \cmark & \cmark & \cmark & \cmark 
& \cmark & \cmark & \cmark 
& Image 
& \cmark \\

\bottomrule
\end{tabular}
}
\caption{Comparison of CCRV-Bench and related benchmarks across causal tasks, constraint mechanisms, input modalities, and evaluation granularity. ``Und.'', ``Pred.'', ``Attr.'', and ``Interv.'' denote causal structure understanding, prediction, attribution, and intervention-outcome prediction, respectively. ``Ground.'' indicates explicit spatial grounding requirements; ``Factual Adversarial Constraint'' indicates verification of false textual premises against visual evidence; and ``Shortcut'' indicates controls targeting shortcut cues. ``Modality'' specifies the input modality, while ``Proc. Eval'' indicates evaluation of intermediate reasoning components beyond the final answer. Checkmarks indicate explicitly evaluated capabilities or implemented mechanisms.}

\label{tab:causal_benchmark_comparison}

\end{table*}

In recent years, VLMs have made significant progress in visual understanding and reasoning tasks, making them potential core decision-making modules in embodied intelligence systems~\cite{driess2023palm}. In real-world physical environments, agents often need to infer causal relationships and predict physical outcomes based solely on single-frame visual observations~\cite{xiang2025aligning}. This process requires the model to identify key entities from static visual inputs, parse their interactive relationships, and conduct structured modeling of potential physical mechanisms. However, if a model relies primarily on language priors or statistical correlations rather than reasoning based on visual evidence, it may produce unreliable predictions in out-of-distribution or novel scenarios~\cite{li2023evaluating}. Therefore, systematically evaluating the visual causal reasoning capabilities of multimodal models in single-image scenarios is crucial for determining their reliability in practical applications.

Recently, some studies have attempted to evaluate multimodal causal reasoning capabilities, as shown in Table~\ref{tab:causal_benchmark_comparison}, but two main limitations remain. First, existing benchmarks typically focus independently on either causal reasoning capabilities~\cite{chen2024cello,komanduri2025causalvlbench,foss2025causalvqa,liu2025causal3d,zhou2024causalbench} or evaluation constraint mechanisms~\cite{zhang2025cfbench,xue2026ccr,wen2024benchmarking}, lacking a systematic combination of the two. Second, existing benchmarks generally lack control mechanisms against shortcut learning. They neither force the model to align causal evidence with task-relevant entity regions nor systematically test robustness to symbolic binding, false-premise verification, and compressed causal expression. Therefore, relying solely on the evaluation results of existing benchmarks may overestimate the actual causal understanding capabilities of these models.

We deconstruct visual causal reasoning into four complementary task dimensions: causal relation discovery, state prediction, causal diagnosis, and intervention-outcome prediction. Motivated by potential reliance on shortcut cues in visual causal reasoning, CCRV-Bench constructs a multi-dimensional constrained evaluation framework. Through entity symbolization, spatial grounding, the factual adversarial constraint, and minimalist output constraints, it probes whether models preserve image-grounded causal performance under different requirements. In summary, the main contributions of this paper are as follows:
\begin{itemize}
\item We propose CCRV-Bench, a constraint-based benchmark for evaluating causal reasoning in multimodal models, featuring a hierarchical evaluation scheme that disentangles distinct components of causal reasoning ability.
\item We develop an orthogonal evaluation framework defined along causal reasoning dimensions and constraint mechanisms, enabling unified quantitative modeling of multi-dimensional causal abilities under diverse adversarial constraints.
\item We show that unconstrained performance does not determine constrained robustness: effective degradation varies across causal tasks and constraint types, exposing failure modes that aggregate causal scores conceal.
\end{itemize}

\begin{figure*}[t]
    \centering
    \includegraphics[width=\linewidth]{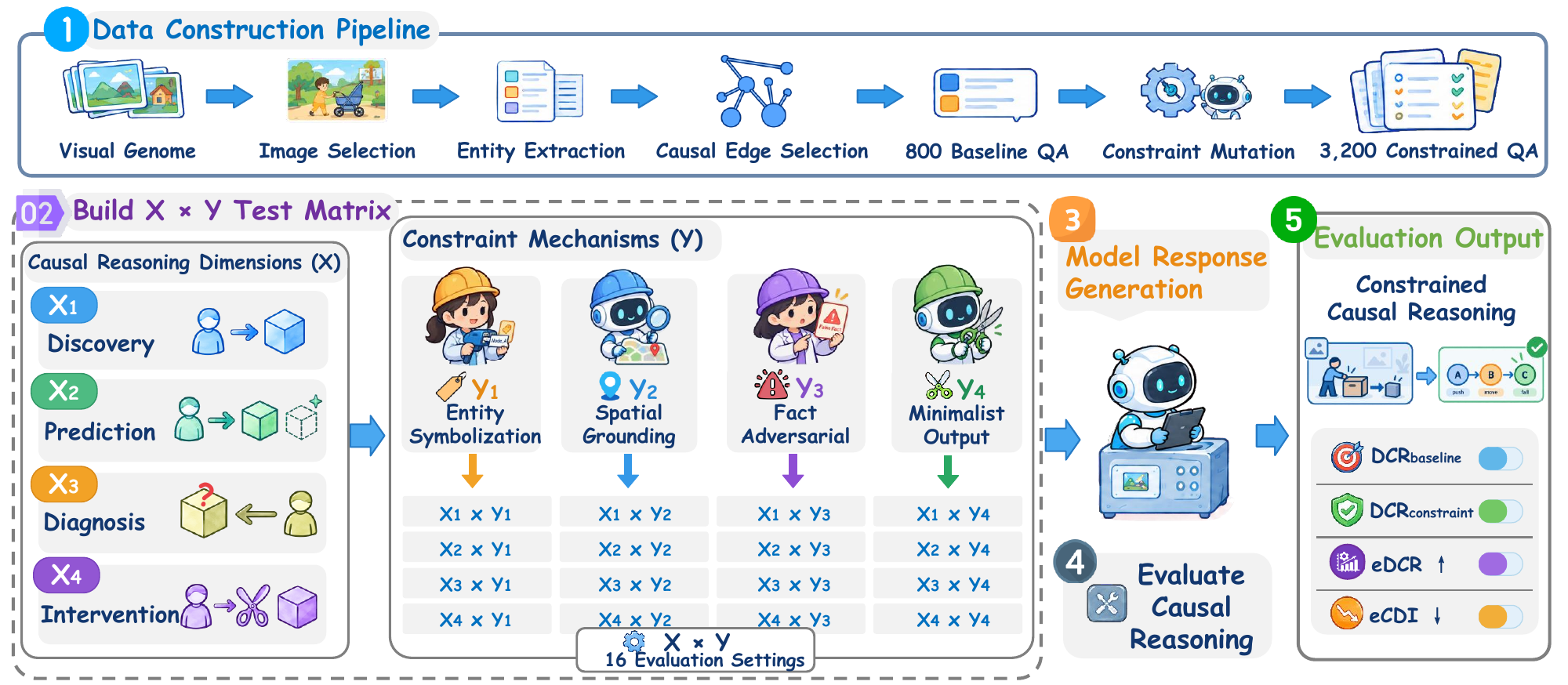}
    \caption{Overview of CCRV-Bench. The top panel presents the data construction pipeline, yielding 800 baseline QA pairs and 3,200 constrained variants. The lower-left panel organizes four causal tasks and four constraint mechanisms into 16 evaluation settings. The right panels show model response generation, evaluation, and the reported metrics.}
    \label{fig:XY}
\end{figure*}
\section{Related Work}
\textbf{Causal Evaluation in VLMs.}
Recent studies have evaluated causal reasoning in VLMs under various structural assumptions. CELLO \citep{chen2024cello} measures models' capabilities in intervention and counterfactual reasoning. CausalVLBench \cite{komanduri2025causalvlbench} investigates multimodal causal understanding in unstructured image-text environments, following Pearl’s causal hierarchy \cite{pearl2009causality, pearl2018book}. Researchers have also explored temporal and physical aspects of causal reasoning. CausalVQA \cite{foss2025causalvqa} evaluates video-based event prediction and hypothetical outcome reasoning, while CAUSAL3D \cite{liu2025causal3d} introduces spatially grounded 3D structures to capture physical interactions often missing in 2D representations. Beyond benchmark construction, diagnostic tools such as structured relevance graphs \cite{pratama2026diagnosing} examine whether models attend to causally relevant image regions during prediction. However, conventional evaluation protocols typically rely on unconstrained natural language prompts, which are prone to inducing shortcut learning \cite{suhail2025shortcut}. Vision-language models often depend on language priors acquired during pre-training \cite{luo2024probing, long2025understanding}, leveraging memorized textual patterns, statistical correlations, or positional cues to generate answers without actively processing visual inputs \cite{buzeta2026seeing}. As a result, standard evaluation settings struggle to disentangle genuine visual causal reasoning from learned statistical regularities.

\textbf{Constrained Evaluation.} Constraint-based evaluation introduces controlled output specifications to regulate model behavior. Existing approaches enforce constraints through compositional structured generation  ~\cite{wen2024benchmarking, zhang2025cfbench} or by verifying predefined conditions and control flows~\cite{xue2026ccr, zhou2023instruction}. A common limitation revealed by such analyses is multimodal hallucination, where models generate plausible yet visually ungrounded content \cite{chen2026survey}. Some studies attempt to mitigate this issue through preference-based selection methods \cite{zhao2025beyond}. These findings suggest that model outputs are not always reliably grounded in visual inputs, even under structured evaluation settings. Moreover, existing approaches tend to focus on specific aspects of model behavior, often examining them in isolation, with limited attention to how constraints influence model performance in tasks requiring structured reasoning.

\section{CCRV-Bench}
This section details the design logic of the constraint-based causal reasoning benchmark. Figure~\ref{fig:XY} illustrates the overall architecture of the benchmark, with its core pipeline encompassing three key modules: data extraction, orthogonal dataset construction, and model evaluation. 
\subsection{Single-Image Causal Modeling}
For each selected image, we model question-relevant visual entities and visually grounded causal relations as a local causal graph \(G=(\mathcal{V},\mathcal{E})\), where nodes denote visual entities and directed edges encode causal interactions supported by visible physical evidence. Here, the nodes $\mathcal{V}$ are visual entities with clear spatial attributes, and the directed edges $\mathcal{E}$ represent the causal interaction mechanisms between entities, encompassing physical contact and spatial constraints. Such interactions affect the state of the target entity and induce subsequent evolution through the action variable $F_{ij}$, which denotes the causal effect of entity $i$ on entity $j$. Based on this underlying logic, this benchmark evaluates four complementary causal task dimensions.

\noindent\textbf{$X_1$: Causal Relation Discovery.} The core task is to parse the causal associations between spatial nodes. Given a single-frame observation, the model must overcome visual noise to accurately locate the initiator and the recipient of an action or state change, denoted as $V_{\text{source}} \xrightarrow{F_{ij}} V_{\text{target}}$. By analyzing microscopic spatial contact or environmental interaction, the model needs to determine the minimal sufficient causal explanation for maintaining or altering the target state. This dimension is responsible for establishing a correct local causal graph within the model, serving as the foundational defense against blind guessing based on global image intuition.

\noindent\textbf{$X_2$: Prediction.}
If $X_1$ examines whether the model can identify a visually grounded causal relation, $X_2$ evaluates whether the model can use that relation to infer its downstream effect. Given a causal interaction observed at time $t_0$, $X_2$ predicts the state evolution of the target entity at time $t_1$. Reliable prediction requires grounding the inference in scene-specific causal evidence. However, shortcut learning may lead models to exploit predictive cues without sufficiently using the relevant causal structure \citep{suhail2025shortcut}. In VLMs, such shortcuts may manifest as over-reliance on generic semantic expectations or stereotypical event patterns, producing plausible outcomes without sufficiently using the visual causal evidence in the current scene.

\noindent\textbf{$X_3$: Causal Diagnosis.} Addressing the need for effect-to-cause deduction in embodied intelligence, when a specific stable state, such as suspension, deformation, or stillness, of an object in a visual scene is given, the model is required to utilize retrospective reasoning to investigate the preconditions. This requires the model to possess rigorous exclusive logical analysis capabilities, enabling it to accurately isolate the most critical causal path from various potential explanations.

\noindent\textbf{$X_4$: Intervention-Outcome Prediction.}
Inspired by Pearl's intervention calculus \citep{pearl2009causality}, this task evaluates the consequences of a specified intervention on a visually grounded causal relation. Given a local removal, release, or state change, the model predicts the resulting state or motion of the target entity. Interventions that disable the relevant causal interaction are denoted as $\mathrm{do}(F_{ij}=0)$.

\subsection{Constraint Mechanism Design}

After establishing the four complementary causal task dimensions, the evaluation framework must address potential shortcut reliance in VLMs. Such shortcuts may allow models to produce plausible responses from multimodal statistical regularities without sufficiently using the relevant visual evidence. Since surface-level success does not necessarily reflect causal reasoning ability \citep{zevcevic2023causal}, we introduce four controlled constraints to reduce explicit entity-name cues, require spatial grounding and false-premise verification, or restrict answer formulation, testing whether models preserve image-grounded causal content under these requirements.

\noindent\textbf{$Y_1$: Entity Symbolization.}
To reduce reliance on lexical identity cues \citep{li2023evaluating}, explicit names of question-relevant entities are replaced with abstract symbols, such as \texttt{Node\_A} and \texttt{Node\_B}, each linked to its corresponding visual region. The model must bind these symbols to the correct visual entities before reasoning about their causal relation. This constraint reduces name-based semantic shortcuts and tests symbol-to-region grounding in causal reasoning.

\noindent\textbf{$Y_2$: Spatial Grounding.}
To address the potential reliance on global visual context for coarse guesses without localizing task-relevant entities \citep{yuksekgonul2022and}, this constraint binds causal explanations to entity regions. The required output consists of one unique coordinate pair $[x,y]$ and a causal explanation. For $X_1$ and $X_3$, the designated region corresponds to the causal-source or partner entity; for $X_2$ and $X_4$, it corresponds to the target entity of the base question. Although the prompt requests image-pixel coordinates, the compatibility scorer checks the coordinate as raw pixels and under valid $[0,1000]$ and $[0,1]$ normalized interpretations mapped to pixels. A coordinate is accepted if at least one valid interpretation falls within the designated entity's bounding box expanded by 7.5\% on each side. This mechanism evaluates whether models can ground their causal explanations in the task-relevant visual region.

\noindent\textbf{$Y_3$: Factual Adversarial Constraint.}
This constraint injects a false textual premise that conflicts with the visible evidence. The model must first reject the false premise and then complete the causal reasoning based on the image-supported causal evidence. By introducing a controlled conflict between text and vision, $Y_3$ tests factual verification and visually grounded causal reasoning under misleading textual cues.

\noindent\textbf{$Y_4$: Minimalist Output.}
In open-ended generation tasks, VLMs may produce lengthy natural-language explanations or complex chains-of-thought that obscure reasoning errors \citep{turpin2023language}. $Y_4$ imposes an item-specific upper bound on response length and restricts punctuation, leaving limited space for verbose reformulation. The model must still express the required causal conclusion within this compressed format, allowing us to assess whether the core causal content is preserved under a tight response budget.

\subsection{Data Construction Pipeline}
\label{sec:data-construction}

The data construction of CCRV-Bench is based on the Visual Genome
(VG) dataset \citep{krishna2017visual}. The system extracts structured
visual entities with bounding boxes $[x,y,w,h]$, categories, and
candidate relations, filtering out pure attribute descriptions and
retaining candidate edges with physical interaction significance.
Human reviewers then checked each local causal-relation specification
used for QA generation against the source image. A relation was retained
as a causal edge when (i) the source and target entities were visually
identifiable, (ii) the visible interaction supported a directional
physical mechanism, and (iii) modifying or removing the interaction
was expected to change the target's state, motion, or stability.
Relations requiring unsupported assumptions or leaving multiple equally
plausible question-relevant causal mechanisms were excluded.

For each of the 100 selected images, GPT-5.1 generated two baseline
question--answer pairs for each of the four causal dimensions from the
verified local causal-relation specifications, yielding
$100 \times 4 \times 2 = 800$ baseline pairs.
GPT-5.1 then instantiated each of the four Y-axis constraints for every baseline pair, producing
3,200 constrained variants organized into an orthogonal $X \times Y$
test matrix. An independent, stratified human audit assessed 160 baseline
pairs (20\% of the baseline set, 40 per causal dimension). Details of
causal-edge filtering, human verification, QA generation, and the audit
protocol and results are provided in Appendix~\ref{sec:baseline-qa-annotation};
the constraint-generation prompts appear in Appendix~\ref{sec:appendix_prompts}.

\subsection{Evaluation Metrics}
To evaluate causal reasoning in visual scenarios, we use the Deep Causal
Reasoning score (DCR) and the Constraint Satisfaction Rate (CSR).
DCR provides a hierarchical assessment of entity grounding, causal
mechanisms, and causal conclusions in model responses based on visual
evidence~\citep{d2022underspecification}.
CSR measures compliance with the assigned Y-axis constraint.

Across the four X-axis tasks, we separately assess each response $i$ for
entity grounding (L1), causal mechanism (L2), and the task-specific causal
conclusion (L3), yielding binary labels
$r_{i,1},r_{i,2},r_{i,3}\in\{0,1\}$.
The cascade sets $r'_{i,1}=r_{i,1}$ and
\begin{equation}
r'_{i,2}=r_{i,1}\land r_{i,2},\qquad
r'_{i,3}=r_{i,1}\land r_{i,2}\land r_{i,3}.
\end{equation}
The response-level score is
\begin{equation}
DCR_i=\frac{r'_{i,1}+r'_{i,2}+r'_{i,3}}{3}.
\end{equation}
Compared with averaging the raw labels, the cascade withholds higher-level
credit when a prerequisite fails. Entity grounding alone is insufficient
for full credit.
We report each model's mean DCR across responses, equivalently the average
of the three cascaded pass rates. DCR therefore allows partial credit.

For the $N$ canonical constrained response slots in a reported aggregate,
let $c_i\in\{0,1\}$ indicate constraint compliance. We define
\begin{equation}
CSR=\frac{1}{N}\sum_{i=1}^{N}c_i,\qquad
\end{equation}
\begin{equation}
eDCR=\frac{1}{N}\sum_{i=1}^{N}DCR_i\,c_i.
\end{equation}
Inspired by safety considerations~\citep{amodei2016concrete}, this hard
gate assigns zero effective credit to noncompliant responses while
retaining graded causal credit for compliant responses.

The CSR rule is instantiated separately for each constraint: $Y_1$ scans for
forbidden real entity names, $Y_2$ requires a unique coordinate whose raw-pixel,
valid $[0,1000]$, or valid $[0,1]$ interpretation falls in the expanded target
entity region, $Y_3$ requires explicit false-premise rejection and a
visual-evidence-based correction, and $Y_4$ checks the prescribed punctuation
and word limit. Therefore, DCR uses the same cascade and numerical scale for
all four $Y$ settings and can be compared descriptively across them. CSR and
eDCR are not interchangeable cross-$Y$ capability scores: their gates have
different semantics and can overlap with the DCR rubric, most strongly for
$Y_3$. Here, ``orthogonal'' $X\times Y$ design refers to the factorial
organization of evaluation cells, not statistical independence of DCR and CSR.

For each constraint, constrained items are linked to their source baseline
items. The main result summaries enumerate the canonical 800 baseline items
and 3,200 constrained variants per model. CDI/eCDI are computed from these fixed-denominator means.
The Constraint Degradation Index and its effective form are
\begin{equation}
CDI=DCR_{\text{baseline}}-DCR_{\text{constraint}},
\end{equation}
\begin{equation}
eCDI=DCR_{\text{baseline}}-eDCR.
\end{equation}
These aggregate differences quantify changes relative to the corresponding
baseline and constrained summaries. CDI captures causal-response degradation, whereas eCDI
also incorporates constraint violations. Positive values indicate
degradation; negative values indicate gains in the corresponding
constrained score. Here, $DCR_{\text{baseline}}$ and
$DCR_{\text{constraint}}$ denote the fixed-denominator means described above.
The formulas are defined on the internal $[0,1]$ scale; DCR and
CSR are reported as percentages, and CDI/eCDI table entries are the
corresponding differences in percentage points.

\section{Experiments}

In this section, we utilize the constructed CCRV-Bench to conduct a comprehensive empirical evaluation of current mainstream VLMs.

\subsection{Experimental Setup}

We evaluate 14 representative VLMs on CCRV-Bench, including 6 closed-source models (GPT-5.5 \citep{leon2025gpt}, Claude Opus 4.7 \citep{anthropic2026claude47}, Gemini-3.1-Pro-Preview \citep{anthropicGemini3.1Pro}, Gemini-3.5-Flash \citep{gemini35flash}, Doubao-Seed-2-Pro \citep{seedance2026seedance20advancingvideo}, and GLM-5V-Turbo \citep{hong2026glm}) and 8 open-weight models (Qwen3.6-35B-A3B \citep{qwen3technicalreport}, Pixtral-12B \citep{agrawal2024pixtral12b}, Qwen2.5-VL-7B \citep{bai2025qwen25vltechnicalreport}, MiniMax-M3 \citep{lai2026minimax}, InternVL3-38B \citep{zhu2025internvl3}, Gemma-3-27B-IT \citep{team2025gemma}, Llama-3.2-11B-Vision \citep{llama32modelcard}, and Mistral-Small-3.2-24B \citep{mistralsmall32modelcard}). More details on these evaluated models can be found in the Appendix Table~\ref{tab:evaluated_models_flat}.
\begin{table*}[!t]
\centering

\resizebox{\textwidth}{!}{
\begin{tabular}{l cccc cccc cccc}
\toprule
\multirow{2}{*}{\textbf{Evaluated Models}} & \multicolumn{4}{c}{\textbf{$\text{DCR}_{\text{baseline}}$}} & \multicolumn{4}{c}{\textbf{$\text{DCR}_{\text{constraint}}$}} & \multirow{2}{*}{\textbf{CSR}} & \multirow{2}{*}{\textbf{eDCR}} & \multirow{2}{*}{\textbf{CDI}} & \multirow{2}{*}{\textbf{eCDI}} \\
\cmidrule(lr){2-5} \cmidrule(lr){6-9}
 & \textbf{L1} & \textbf{L2} & \textbf{L3} & \textbf{DCR} & \textbf{L1} & \textbf{L2} & \textbf{L3} & \textbf{DCR} & & & & \\
\midrule
\textbf{GPT-5.5} & 92.6 & 89.3 & \textbf{86.1} & 89.3 & \textbf{86.9} & \textbf{79.2} & \textbf{70.8} & \textbf{79.0} & \textbf{95.0} & \textbf{76.9} & 10.4 & \textbf{12.4} \\
\textbf{Claude-Opus-4.7} & \textbf{95.8} & \underline{91.0} & \underline{85.3} & \textbf{90.7} & \underline{84.7} & \underline{76.8} & \underline{70.2} & \underline{77.2} & \underline{93.0} & \underline{74.5} & 13.5 & 16.1 \\
\textbf{Doubao-Seed-2-Pro} & \underline{95.4} & \textbf{91.9} & 82.3 & \underline{89.8} & 70.2 & 61.6 & 55.8 & 62.6 & 91.9 & 60.0 & 27.3 & 29.8 \\
\textbf{Gemini-3.1-Pro-Preview} & 93.9 & 88.3 & 79.6 & 87.3 & 71.9 & 59.6 & 48.5 & 60.0 & 92.4 & 58.6 & 27.3 & 28.6 \\
\textbf{Gemini-3.5-Flash} & 94.9 & 90.5 & 84.0 & \underline{89.8} & 69.2 & 59.8 & 51.3 & 60.1 & 86.3 & 56.2 & 29.7 & 33.6 \\
\textbf{GLM-5V-Turbo} & 86.3 & 78.6 & 67.6 & 77.5 & 71.3 & 61.8 & 53.6 & 62.2 & 92.9 & 60.4 & 15.3 & 17.1 \\
\textbf{Pixtral-12B} & 86.8 & 78.5 & 66.0 & 77.1 & 64.2 & 51.5 & 41.8 & 52.5 & 71.6 & 43.7 & 24.6 & 33.4 \\
\textbf{Qwen2.5-VL-7B} & 73.4 & 62.1 & 48.8 & 61.4 & 69.3 & 49.9 & 37.7 & 52.3 & 77.9 & 45.2 & \underline{9.1} & 16.3 \\
\textbf{Qwen3.6-35B-A3B} & 80.5 & 72.6 & 64.5 & 72.5 & 64.5 & 50.3 & 40.5 & 51.8 & 92.8 & 50.7 & 20.8 & 21.8 \\
\textbf{InternVL3-38B} & 79.4 & 68.3 & 56.8 & 68.1 & 68.0 & 53.4 & 45.0 & 55.5 & 86.7 & 52.0 & 12.7 & 16.2 \\
\textbf{Gemma-3-27B-IT} & 77.5 & 65.8 & 53.1 & 65.5 & 63.3 & 48.3 & 40.8 & 50.8 & 79.5 & 44.9 & 14.7 & 20.6 \\
\textbf{Llama-3.2-11B-Vision} & 62.0 & 48.0 & 35.6 & 48.5 & 61.4 & 43.6 & 34.7 & 46.6 & 52.8 & 32.8 & \textbf{2.0} & \underline{15.7} \\
\textbf{Mistral-Small-3.2-24B} & 87.9 & 77.0 & 64.4 & 76.4 & 72.5 & 57.1 & 47.1 & 58.9 & 82.3 & 52.4 & 17.5 & 24.0 \\
\textbf{MiniMax-M3} & 93.8 & 89.3 & 81.5 & 88.2 & 80.9 & 73.7 & 67.4 & 74.0 & 38.2 & 34.2 & 14.2 & 54.0 \\

\bottomrule
\end{tabular}
}
\caption{Main evaluation results of 14 VLMs on CCRV-Bench. L1--L3 report cascaded pass rates, and DCR is their mean. DCR, CSR, and eDCR are reported as percentages; CDI and eCDI are reported in percentage points. The best and second-best scores are highlighted in bold and underlined, respectively.}
\label{tab:ccrv_results}
\end{table*}

\subsection{Results}
\begin{table*}[!t]
\centering

\resizebox{\textwidth}{!}{
\begin{tabular}{l cccc cccc cccc}
\toprule
\raisebox{-1.5ex}{\textbf{Evaluated Models}} & \multicolumn{4}{c}{\textbf{$\text{DCR}_{\text{baseline}}$ }} & \multicolumn{4}{c}{\textbf{$\text{DCR}_{\text{constraint}}$ }} & \multicolumn{4}{c}{\textbf{eCDI}} \\
\cmidrule(lr){2-5} \cmidrule(lr){6-9} \cmidrule(lr){10-13}
& $X_1$ & $X_2$ & $X_3$ & $X_4$ & $X_1$ & $X_2$ & $X_3$ & $X_4$ & $X_1$ & $X_2$ & $X_3$ & $X_4$ \\
\midrule
\textbf{GPT-5.5} & 82.3 & \textbf{92.7} & 88.2 & \textbf{94.2} & \textbf{82.3} & \textbf{78.4} & \textbf{78.6} & \underline{76.6} & \textbf{1.8} & 16.5 & 11.8 & 19.5 \\
\textbf{Claude-Opus-4.7} & \underline{89.7} & \underline{91.0} & \underline{89.3} & 92.7 & \underline{77.0} & \underline{76.8} & \underline{76.9} & \textbf{78.1} & 15.0 & 17.0 & 15.8 & \underline{16.7} \\
\textbf{Doubao-Seed-2-Pro} & \textbf{91.5} & 90.0 & 87.3 & 90.5 & 67.1 & 61.3 & 63.8 & 58.0 & 27.3 & 30.8 & 27.4 & 33.7 \\
\textbf{Gemini-3.1-Pro-Preview} & 85.7 & 85.8 & 88.0 & 89.5 & 65.4 & 56.6 & 60.4 & 57.5 & 22.0 & 29.8 & 29.8 & 33.0 \\
\textbf{Gemini-3.5-Flash} & 86.0 & 88.5 & \textbf{91.7} & \underline{93.0} & 64.3 & 58.3 & 60.9 & 56.9 & 25.6 & 34.5 & 34.4 & 40.1 \\
\textbf{GLM-5V-Turbo} & 71.5 & 76.5 & 79.3 & 82.7 & 64.1 & 63.3 & 62.8 & 58.6 & \underline{10.0} & \underline{14.9} & 18.9 & 24.4 \\
\textbf{Pixtral-12B} & 70.8 & 77.8 & 75.0 & 84.7 & 48.6 & 51.9 & 52.9 & 56.6 & 29.6 & 34.5 & 32.5 & 37.0 \\
\textbf{Qwen2.5-VL-7B} & 57.3 & 73.5 & 47.3 & 67.5 & 51.2 & 55.0 & 47.8 & 55.2 & 15.3 & 23.5 & \textbf{9.3} & 17.0 \\
\textbf{Qwen3.6-35B-A3B} & 73.8 & 72.5 & 71.8 & 72.0 & 54.9 & 52.4 & 50.8 & 49.0 & 20.3 & 20.9 & 22.2 & 24.0 \\
\textbf{InternVL3-38B} & 65.8 & 66.8 & 63.7 & 76.2 & 56.7 & 54.6 & 57.9 & 52.6 & 12.5 & \underline{14.9} & \underline{10.1} & 27.1 \\
\textbf{Gemma-3-27B-IT} & 57.5 & 72.5 & 59.2 & 72.7 & 53.4 & 51.3 & 50.4 & 48.1 & 10.5 & 26.7 & 15.9 & 29.2 \\
\textbf{Llama-3.2-11B-Vision} & 48.8 & 40.2 & 52.3 & 52.8 & 45.2 & 47.8 & 43.9 & 49.5 & 18.5 & \textbf{5.8} & 22.3 & \textbf{16.2} \\
\textbf{Mistral-Small-3.2-24B} & 72.7 & 80.3 & 68.0 & 84.7 & 58.6 & 59.8 & 57.0 & 60.4 & 19.7 & 25.7 & 17.5 & 33.3 \\
\textbf{MiniMax-M3} & 86.3 & 89.7 & 84.7 & 92.0 & 73.8 & 72.9 & 73.8 & 75.4 & 50.1 & 57.5 & 49.3 & 59.2 \\
\bottomrule
\end{tabular}
}
\caption{Detailed cognitive performance breakdown across four causal task dimensions ($X_1$: Discovery, $X_2$: Prediction, $X_3$: Diagnosis, $X_4$: Intervention-Outcome Prediction). DCR values are percentages and eCDI values are percentage-point drops (the internal differences are multiplied by 100). Best results are \textbf{bolded}, and second-best are \underline{underlined}.}
\label{tab:x_dimensions_trend}
\end{table*}

Table~\ref{tab:ccrv_results} summarizes the main aggregate and stage-wise results of the 14 models evaluated on CCRV-Bench. Taken together, these results characterize model performance on unconstrained causal questions and under explicit task constraints.

\noindent\textbf{Performance Differences Under Constraints:}
Gemini-3.5-Flash slightly outperforms GPT-5.5 at baseline (89.8\% vs.\ 89.3\%), suggesting comparable correctness on unconstrained causal questions. Under constraints, however, Gemini loses 29.7 DCR points, nearly three times GPT's 10.4-point loss. High scores on unconstrained benchmarks are therefore insufficient to establish reliable causal reasoning under the additional requirements of practical tasks. Beyond differences between models, the stage-wise evaluation across all 14 models reveals difficulties at multiple stages of causal reasoning. Across the 14 models, the average L1 score is 85.7\% in the baseline setting and 71.3\% under constraints. The L1--L2 gap also widens from 7.8 to 12.3 points. Under the cascaded scoring scheme, this gap represents responses that pass entity grounding but fail the causal mechanism criterion. Complete-chain correctness (L3) falls from 68.3\% to 50.4\%; about half of the constrained responses therefore satisfy entity grounding, causal mechanism, and causal conclusion simultaneously. Under explicit constraints, models thus struggle to connect the relevant entities with the physical mechanisms and causal conclusions needed for a complete, correct causal answer.

\noindent\textbf{Different Failure Patterns:}
For 12 of the 14 models, the DCR decline under constraints is larger than the score lost through constraint violations. MiniMax-M3 and GLM-5V-Turbo illustrate how causal correctness and constraint following can diverge. MiniMax achieves a relatively high constrained DCR (74.0\%) but a low CSR (38.2\%): many responses receive DCR credit but fail to satisfy the constraints, reducing its eDCR to 34.2\%. GLM has a lower constrained DCR (62.2\%) but a much higher CSR (92.9\%), achieving an eDCR of 60.4\%. MiniMax therefore has the higher constrained DCR, whereas GLM performs better when both causal correctness and constraint following are required. Llama-3.2-11B-Vision shows another limitation. Its CDI is only 2.0 points, but its baseline DCR is also low (48.5\%), so the small decline alone does not establish strong performance under constraints. For Llama-3.2-11B-Vision, its constrained DCR of 46.6\% falls by a further 13.8 points once constraint satisfaction is applied, leaving its eDCR at 32.8\%. These results reveal different weaknesses in constrained causal reasoning: relatively accurate causal answers can coexist with poor constraint following, while a small decline from baseline can coexist with low overall correctness.

\section{Additional Analysis}
\begin{table*}[!t]
\centering

\resizebox{0.9\textwidth}{!}{
\begin{tabular}{l cccc cccc cccc}
\toprule
\raisebox{-1.5ex}{\textbf{Evaluated Models}} & \raisebox{-1.5ex}{\textbf{$\text{DCR}_{\text{baseline}}$}} & \multicolumn{4}{c}{\textbf{$\text{DCR}_{\text{constraint}}$ }} & \multicolumn{4}{c}{\textbf{eCDI}} \\
\cmidrule(lr){3-6} \cmidrule(lr){7-10}
& & $Y_1$ & $Y_2$ & $Y_3$ & $Y_4$ & $Y_1$ & $Y_2$ & $Y_3$ & $Y_4$ \\
\midrule
\textbf{GPT-5.5} & 89.3 & \textbf{77.0} & \textbf{75.5} & \textbf{97.7} & \textbf{65.7} & \underline{12.5} & \textbf{21.5} & -8.4 & 23.9 \\
\textbf{Claude-Opus-4.7} & \textbf{90.7} & \underline{75.5} & \underline{74.8} & 95.2 & 63.3 & 17.2 & \underline{23.8} & -4.2 & 27.8 \\
\textbf{Doubao-Seed-2-Pro} & \underline{89.8} & 72.4 & 32.4 & 92.3 & 53.1 & 22.7 & 62.2 & -2.5 & 36.7 \\
\textbf{Gemini-3.1-Pro-Preview} & 87.3 & 48.3 & 32.3 & 94.7 & \underline{64.8} & 39.0 & 60.1 & -7.0 & 22.5 \\
\textbf{Gemini-3.5-Flash} & \underline{89.8} & 48.6 & 37.6 & \underline{95.3} & 58.8 & 47.0 & 61.7 & -5.2 & 31.0 \\
\textbf{GLM-5V-Turbo} & 77.5 & 58.8 & 33.5 & 92.9 & 63.7 & 20.4 & 49.2 & -15.3 & \textbf{13.9} \\
\textbf{Pixtral-12B} & 77.1 & 50.9 & 31.2 & 86.5 & 41.5 & 27.3 & 59.8 & -9.0 & 55.5 \\
\textbf{Qwen2.5-VL-7B} & 61.4 & 43.5 & 46.2 & 71.6 & 47.9 & 21.9 & 28.0 & -7.1 & 22.2 \\
\textbf{Qwen3.6-35B-A3B} & 72.5 & 43.6 & 20.2 & 90.4 & 53.0 & 29.6 & 55.2 & -17.2 & \underline{19.8} \\
\textbf{InternVL3-38B} & 68.1 & 48.1 & 32.0 & 90.7 & 51.0 & 20.3 & 45.7 & -22.4 & 21.1 \\
\textbf{Gemma-3-27B-IT} & 65.5 & 26.7 & 37.0 & 91.5 & 47.9 & 39.5 & 42.5 & \underline{-25.9} & 26.2 \\
\textbf{Llama-3.2-11B-Vision} & 48.5 & 43.8 & 18.7 & 81.5 & 42.4 & \textbf{7.8} & 41.0 & \textbf{-31.4} & 45.4 \\
\textbf{Mistral-Small-3.2-24B} & 76.4 & 51.0 & 45.2 & 89.3 & 50.3 & 26.3 & 48.5 & -12.8 & 34.0 \\
\textbf{MiniMax-M3} & 88.2 & 73.7 & 70.8 & 92.9 & 58.6 & 84.4 & 48.1 & -4.7 & 88.2 \\

\bottomrule
\end{tabular}
}
\caption{Detailed performance breakdown across four specific adversarial constraints ($Y_1$: Entity Symbolization, $Y_2$: Entity-Region Spatial Grounding, $Y_3$: Factual Adversarial Constraint, $Y_4$: Minimalist Output). DCR values are percentages and eCDI values are percentage-point drops (the internal differences are multiplied by 100). Best results are \textbf{bolded}, and second-best are \underline{underlined}.}
\label{tab:y_constraints_trend}
\end{table*}

\noindent\textbf{X Dimension: From Causal Discovery to Intervention-Outcome Prediction.}
Table~\ref{tab:x_dimensions_trend} covers the full causal reasoning process from discovering a causal relation ($X_1$), to predicting its consequence ($X_2$), diagnosing the cause of an observed outcome ($X_3$), and predicting the outcome of a specified intervention ($X_4$). The results show that models are relatively stable when identifying causal structure already present in the image, but become less reliable when that structure must be used to infer unseen consequences or altered outcomes. Mean DCR drops by 12.7 points on discovery ($X_1$), increases to 18.4 points on prediction ($X_2$), becomes smaller on diagnosis ($X_3$, 14.9 points), and reaches its largest decline on intervention-outcome prediction ($X_4$, 22.3 points). This pattern suggests that the main difficulty is not simply recognizing causal relations, but using them consistently across forward prediction, backward diagnosis, and especially predicting the consequences of an intervention.

GPT-5.5 makes this gap particularly clear: its discovery score remains unchanged at 82.3\%, whereas its intervention score drops from 94.2\% to 76.6\%. At the same time, degradation must be interpreted relative to the starting capability. Llama-3.2-11B-Vision shows a relatively small discovery decline, from 48.8\% to 45.2\%, but starts from a low baseline, while Pixtral-12B drops from 70.8\% to 48.6\%. Thus, the magnitude of the decline alone is insufficient to establish robustness: a small decline may reflect either stable performance from a strong starting point or persistently limited performance from a weak one. Taken together, the X-wise results show that current models are better at recognizing causal structure than reliably using it throughout the full causal reasoning chain, particularly when the task requires predicting the consequences of an intervention.

\noindent\textbf{Y Dimension: Constraint-Specific Effects.}
Table~\ref{tab:y_constraints_trend} reveals sharply different effects across the four constraints. Spatial grounding ($Y_2$) causes the strongest degradation, with the mean constrained DCR falling to 41.9\%, and yields the lowest constrained DCR for 9 of the 14 models. This indicates that when models must not only identify a causal relation but also ground it to the task-relevant entity region, maintaining causal correctness becomes substantially more difficult. $Y_2$ evaluates this entity-region grounding rather than precise contact-point localization. This matters because downstream decisions depend on whether the inferred relation is grounded in the correct part of the observed scene. Entity symbolization ($Y_1$) and minimalist output ($Y_4$) produce more moderate but still substantial degradation, suggesting that causal performance also becomes less stable when entity-name cues are removed or when the same causal relation must be expressed under strict output constraints.

The factual adversarial constraint ($Y_3$) shows the opposite pattern. All 14 models exceed their own baseline DCR under $Y_3$, producing a mean DCR of 90.2\% and a negative mean eCDI of $-12.4$ points. Higher task-conditioned scores are observed under explicit false-premise verification, which may direct attention to relevant visual evidence, although changes in question content and scoring criteria may also contribute. We further analyze this effect in Appendix~\ref{sec:appendix_y3_analysis}. Overall, the Y-axis results show that causal correctness under standard prompting does not necessarily carry over when the same reasoning must be grounded, reformulated, or checked against misleading information. These requirements are not merely additional difficulty; they reflect conditions that causal reasoning must satisfy when it is used beyond unconstrained question answering.

\subsection{Independent Human Validation}
Three annotators independently conducted a blinded, stratified audit of 280 responses, with one response sampled from each combination of 14 models, five settings, and four causal tasks. Fleiss' $\kappa$ is 0.7143, 0.7848, and 0.8365 for L1--L3, respectively, and 0.7415 for full-chain pass. Human-majority agreement with GPT-5.1 is 81.07\%, 83.93\%, and 87.14\% for L1--L3, respectively, and 77.50\% for full-chain pass. On the audited subset, human-majority and GPT-5.1 scoring agree on the direction of score differences from Baseline across all four constraints, although their magnitudes differ. Detailed protocols, agreement statistics, setting-level diagnostics, and the separate 56-response $Y_3$ CSR audit are provided in Appendix~\ref{app:human_agreement}.
\section{Conclusion}
This paper introduces CCRV-Bench, designed to evaluate the visual causal reasoning capabilities of VLMs under constraints. We construct an orthogonal evaluation framework encompassing causal structure discovery, causal prediction, causal attribution, and intervention-outcome prediction, conducting rigorous evaluations of mainstream models under constrained conditions. The results indicate that strong performance on unconstrained causal questions does not consistently translate into strong performance under explicit constraints. Performance varies across models and constraint types, revealing difficulties in maintaining causal correctness while satisfying additional task requirements. CCRV-Bench reveals limitations in the constrained causal performance of current multimodal models and emphasizes the value of constrained evaluation in characterizing image-grounded causal reasoning. This provides a diagnostic foundation for assessing and developing multimodal and embodied intelligence systems equipped with reliable causal reasoning capabilities.

\section*{Limitations}
While CCRV-Bench provides a standardized framework for diagnosing the genuine causal cognitive capabilities of VLMs, this study still presents several limitations. First, our benchmark focuses exclusively on single-frame static scenarios. While this effectively isolates the model's ability to infer structural causality from spatial snapshots, real-world embodied agents operate in continuous temporal environments, where dynamic causal reasoning based on videos or 3D interactions is equally crucial. A paired-image extension could use the first image as the pre-condition and the second as the post-action or post-intervention outcome. For videos, the $X_1$--$X_4$ tasks could be mapped to causal-event identification, future-state prediction, diagnosis of an earlier cause, and prediction of an intervention outcome, while retaining the same L1/L2/L3 DCR structure over time. Second, the physical scenes in our dataset are primarily sourced from Visual Genome, which predominantly covers everyday commonsense interactions. Consequently, it may not comprehensively evaluate a model's causal reasoning capabilities in highly specialized or complex physical domains. Third, despite rigorous manual verification, utilizing an LLM-in-the-loop pipeline for constraint mutation might introduce subtle linguistic biases or prompt structures, inadvertently favoring models with similar alignment paradigms. Finally, our constraints primarily target semantic, spatial, and logical shortcuts. By extending this framework to temporal modalities and broader physical domains, we hope to continue advancing the development of reliable embodied AI systems.

\bibliography{custom}

@inproceedings{chen2024cello,
  title={Cello: Causal evaluation of large vision-language models},
  author={Chen, Meiqi and Peng, Bo and Zhang, Yan and Lu, Chaochao},
  booktitle={Proceedings of the 2024 Conference on Empirical Methods in Natural Language Processing},
  pages={22353--22374},
  year={2024}
}

@inproceedings{komanduri2025causalvlbench,
  title={Causalvlbench: Benchmarking visual causal reasoning in large vision-language models},
  author={Komanduri, Aneesh and Bhaila, Karuna and Wu, Xintao},
  booktitle={Proceedings of the 2025 Conference on Empirical Methods in Natural Language Processing},
  pages={30648--30668},
  year={2025}
}

@article{li2026incompressible,
  title={Incompressible knowledge probes: Estimating black-box llm parameter counts via factual capacity},
  author={Li, Bojie},
  journal={arXiv preprint arXiv:2604.24827},
  year={2026}
}

@article{foss2025causalvqa,
  title={Causalvqa: A physically grounded causal reasoning benchmark for video models},
  author={Foss, Aaron and Evans, Chloe and Mitts, Sasha and Sinha, Koustuv and Rizvi, Ammar and Kao, Justine T},
  journal={arXiv preprint arXiv:2506.09943},
  year={2025}
}

@article{liu2025causal3d,
  title={Causal3d: A comprehensive benchmark for causal learning from visual data},
  author={Liu, Disheng and Qiao, Yiran and Liu, Wuche and Lu, Yiren and Zhou, Yunlai and Liang, Tuo and Yin, Yu and Ma, Jing},
  journal={arXiv preprint arXiv:2503.04852},
  year={2025}
}

@article{pratama2026diagnosing,
  title={Diagnosing Causal Reasoning in Vision-Language Models via Structured Relevance Graphs},
  author={Pratama, Dhita Putri and Han, Soyeon Caren and Ding, Yihao},
  journal={arXiv preprint arXiv:2602.20878},
  year={2026}
}

@article{suhail2025shortcut,
  title={Shortcut learning susceptibility in vision classifiers},
  author={Suhail, Pirzada and Goel, Vrinda and Sethi, Amit},
  journal={arXiv preprint arXiv:2502.09150},
  year={2025}
}

@article{luo2024probing,
  title={Probing visual language priors in vlms},
  author={Luo, Tiange and Cao, Ang and Lee, Gunhee and Johnson, Justin and Lee, Honglak},
  journal={arXiv preprint arXiv:2501.00569},
  year={2024}
}

@article{long2025understanding,
  title={Understanding language prior of lvlms by contrasting chain-of-embedding},
  author={Long, Lin and Oh, Changdae and Park, Seongheon and Li, Sharon},
  journal={arXiv preprint arXiv:2509.23050},
  year={2025}
}

@article{buzeta2026seeing,
  title={Seeing to Generalize: How Visual Data Corrects Binding Shortcuts},
  author={Buzeta, Nicolas and del Rio, Felipe and Hinostroza, Cristian and Parra, Denis and Lobel, Hans and Icarte, Rodrigo Toro},
  journal={arXiv preprint arXiv:2602.15183},
  year={2026}
}

@article{wen2024benchmarking,
  title={Benchmarking complex instruction-following with multiple constraints composition},
  author={Wen, Bosi and Ke, Pei and Gu, Xiaotao and Wu, Lindong and Huang, Hao and Zhou, Jinfeng and Li, Wenchuang and Hu, Binxin and Gao, Wendy and Xu, Jiaxin and others},
  journal={Advances in Neural Information Processing Systems},
  volume={37},
  pages={137610--137645},
  year={2024}
}

@inproceedings{zhang2025cfbench,
  title={Cfbench: A comprehensive constraints-following benchmark for llms},
  author={Zhang, Tao and Zhu, Chenglin and Shen, Yanjun and Luo, Wenjing and Zhang, Yan and Liang, Hao and Yang, Fan and Lin, Mingan and Qiao, Yujing and Chen, Weipeng and others},
  booktitle={Proceedings of the 63rd Annual Meeting of the Association for Computational Linguistics (Volume 1: Long Papers)},
  pages={32926--32944},
  year={2025}
}

@article{xue2026ccr,
  title={CCR-Bench: A Comprehensive Benchmark for Evaluating LLMs on Complex Constraints, Control Flows, and Real-World Cases},
  author={Xue, Xiaona and Huang, Yiqiao and Li, Jiacheng and Zheng, Yuanhang and Miao, Huiqi and Ma, Yunfei and Liu, Rui and Sun, Xinbao and Liu, Minglu and Meng, Fanyu and others},
  journal={arXiv preprint arXiv:2603.07886},
  year={2026}
}

@article{zhou2023instruction,
  title={Instruction-following evaluation for large language models},
  author={Zhou, Jeffrey and Lu, Tianjian and Mishra, Swaroop and Brahma, Siddhartha and Basu, Sujoy and Luan, Yi and Zhou, Denny and Hou, Le},
  journal={arXiv preprint arXiv:2311.07911},
  year={2023}
}

@article{chen2026survey,
  title={A survey of multimodal hallucination evaluation and detection},
  author={Chen, Zhiyuan and Min, Yuecong and Zhang, Jie and Yan, Bei and Wang, Jiahao and Wang, Xiaozhen and Shan, Shiguang},
  journal={International Journal of Computer Vision},
  volume={134},
  number={3},
  pages={131},
  year={2026},
  publisher={Springer}
}

@inproceedings{zhao2025beyond,
  title={Beyond Multimodal Hallucinations: Enhancing LVLMs through Hallucination-Aware Direct Preference Optimization},
  author={Zhao, Zhiyuan and Wang, Bin and Ouyang, Linke and Dong, Xiaoyi and Wang, Jiaqi and He, Conghui},
  booktitle={2025 IEEE International Conference on Multimedia and Expo (ICME)},
  pages={1--6},
  year={2025},
  organization={IEEE}
}

@book{pearl2009causality,
  title={Causality},
  author={Pearl, Judea},
  year={2009},
  publisher={Cambridge university press}
}

@book{pearl2018book,
  title={The book of why: The new science of cause and effect},
  author={Pearl, Judea},
  year={2018},
  publisher={Basic Books}
}

@article{zevcevic2023causal,
  title={Causal parrots: Large language models may talk causality but are not causal},
  author={Ze{\v{c}}evi{\'c}, Matej and Willig, Moritz and Dhami, Devendra Singh and Kersting, Kristian},
  journal={arXiv preprint arXiv:2308.13067},
  year={2023}
}

@inproceedings{li2023evaluating,
  title={Evaluating object hallucination in large vision-language models},
  author={Li, Yifan and Du, Yifan and Zhou, Kun and Wang, Jinpeng and Zhao, Xin and Wen, Ji-Rong},
  booktitle={Proceedings of the 2023 conference on empirical methods in natural language processing},
  pages={292--305},
  year={2023}
}

@article{yuksekgonul2022and,
  title={When and why vision-language models behave like bags-of-words, and what to do about it?},
  author={Yuksekgonul, Mert and Bianchi, Federico and Kalluri, Pratyusha and Jurafsky, Dan and Zou, James},
  journal={arXiv preprint arXiv:2210.01936},
  year={2022}
}

@article{turpin2023language,
  title={Language models don't always say what they think: Unfaithful explanations in chain-of-thought prompting},
  author={Turpin, Miles and Michael, Julian and Perez, Ethan and Bowman, Samuel},
  journal={Advances in Neural Information Processing Systems},
  volume={36},
  pages={74952--74965},
  year={2023}
}

@article{krishna2017visual,
  title={Visual genome: Connecting language and vision using crowdsourced dense image annotations},
  author={Krishna, Ranjay and Zhu, Yuke and Groth, Oliver and Johnson, Justin and Hata, Kenji and Kravitz, Joshua and Chen, Stephanie and Kalantidis, Yannis and Li, Li-Jia and Shamma, David A and others},
  journal={International journal of computer vision},
  volume={123},
  number={1},
  pages={32--73},
  year={2017},
  publisher={Springer}
}

@article{d2022underspecification,
  title={Underspecification presents challenges for credibility in modern machine learning},
  author={D'Amour, Alexander and Heller, Katherine and Moldovan, Dan and Adlam, Ben and Alipanahi, Babak and Beutel, Alex and Chen, Christina and Deaton, Jonathan and Eisenstein, Jacob and Hoffman, Matthew D and others},
  journal={Journal of Machine Learning Research},
  volume={23},
  number={226},
  pages={1--61},
  year={2022}
}

@article{amodei2016concrete,
  title={Concrete problems in AI safety},
  author={Amodei, Dario and Olah, Chris and Steinhardt, Jacob and Christiano, Paul and Schulman, John and Man{\'e}, Dan},
  journal={arXiv preprint arXiv:1606.06565},
  year={2016}
}

@article{leon2025gpt,
  title={GPT-5 and open-weight large language models: Advances in reasoning, transparency, and control},
  author={Leon, Maikel},
  journal={Information Systems},
  pages={102620},
  year={2025},
  publisher={Elsevier}
}

@misc{anthropic2026claude47,
  title={Introducing Claude Opus 4.7},
  author={Anthropic},
  url={https://www.anthropic.com/news/claude-opus-4-7},
  year={2026-04-16},
  note={Accessed: 2026-05-22} 
}

@misc{seedance2026seedance20advancingvideo,
  author = {{ByteDance Seed}},
  title = {{Seed2.0}},
  year = {2026},
  url = {https://seed.bytedance.com/zh/seed2},
  note = {Accessed: 2026-08-24}
}

@misc{anthropicGemini3.1Pro,
  title = {Model Evaluation – Approach, Methodology \& Results Gemini 3.1 Pro},
  author = {Google DeepMind},
  year = {2026},
  month = {Feb},
  url ={https://storage.googleapis.com/deepmind-media/gemini/gemini_3-1_pro_model_evaluation.pdf},
  note = {Accessed: 2026-05-22}
}

@article{hong2026glm,
  title={GLM-5V-Turbo: Toward a Native Foundation Model for Multimodal Agents},
  author={Hong, Wenyi and Gu, Xiaotao and Pan, Ziyang and Yang, Zhen and Wang, Yuting and Wang, Yue and Yue, Yuanchang and Wang, Yu and Wang, Yanling and Wang, Yan and others},
  journal={arXiv preprint arXiv:2604.26752},
  year={2026}
}

@misc{qwen3technicalreport,
  author = {Qwen Team},
  title = {Qwen3.6-35B-A3B: Agentic Coding Power, Now Open to All},
  year = {2026},
  url={https://qwen.ai/blog?id=qwen3.6-35b-a3b},
  note = {Accessed: 2026-05-22}
}

@article{zhu2025internvl3,
  title={Internvl3: Exploring advanced training and test-time recipes for open-source multimodal models},
  author={Zhu, Jinguo and Wang, Weiyun and Chen, Zhe and Liu, Zhaoyang and Ye, Shenglong and Gu, Lixin and Tian, Hao and Duan, Yuchen and Su, Weijie and Shao, Jie and others},
  journal={arXiv preprint arXiv:2504.10479},
  year={2025}
}

@article{team2025gemma,
  title={Gemma 3 technical report},
  author={Team, Gemma and Kamath, Aishwarya and Ferret, Johan and Pathak, Shreya and Vieillard, Nino and Merhej, Ramona and Perrin, Sarah and Matejovicova, Tatiana and Ram{\'e}, Alexandre and Rivi{\`e}re, Morgane and others},
  journal={arXiv preprint arXiv:2503.19786},
  year={2025}
}

@misc{llama32modelcard,
  author = {{Meta}},
  title = {Model Cards \& Prompt Formats: {Llama 3.2}},
  year = {2024},
  url = {https://developer.meta.com/ai/docs/model-cards-and-prompt-formats/llama3_2/},
  note = {Accessed: 2026-05-22}
}

@misc{mistralsmall32modelcard,
  author = {{Mistral AI}},
  title = {{Mistral-Small-3.2-24B-Instruct-2506}},
  year = {2025},
  url = {https://huggingface.co/mistralai/Mistral-Small-3.2-24B-Instruct-2506},
  note = {Accessed: 2026-05-22}
}

@article{lai2026minimax,
  title={Minimax sparse attention},
  author={Lai, Xunhao and Xu, Weiqi and Yang, Yufeng and Chen, Qiaorui and Xu, Yang and Zeng, Lunbin and Li, Xiaolong and Sun, Haohai and Zhu, Haichao and Zhang, Vito and others},
  journal={arXiv preprint arXiv:2606.13392},
  year={2026}
}

@article{bai2025qwen25vltechnicalreport,
  title={Qwen2.5-VL technical report. CoRR abs/2502.13923 (2025)},
  author={Bai, Shuai and Chen, Keqin and Liu, Xuejing and Wang, Jialin and Ge, Wenbin and Song, Sibo and Dang, Kai and Wang, Peng and Wang, Shijie and Tang, Jun and others},
  journal={arXiv preprint arXiv:2502.13923},
  year={2025}
}

@misc{gemini35flash,
  title={Gemini 3.5 Flash: The new leader in intelligence versus speed},
  author={Google DeepMind},
  url={https://artificialanalysis.ai/articles/gemini-3-5-flash-everything-you-need-to-know},
  year={2026},
  note={Accessed: 2026-05-22} 
}

@article{agrawal2024pixtral12b,
  title={Pixtral 12B},
  author={Agrawal, Pravesh and Antoniak, Szymon and Hanna, Emma Bou and Bout, Baptiste and Chaplot, Devendra and Chudnovsky, Jessica and Costa, Diogo and De Monicault, Baudouin and Garg, Saurabh and Gervet, Theophile and others},
  journal={arXiv preprint arXiv:2410.07073},
  year={2024}
}

@article{zhou2024causalbench,
  title={Causalbench: A comprehensive benchmark for causal learning capability of llms},
  author={Zhou, Yu and Wu, Xingyu and Huang, Beicheng and Wu, Jibin and Feng, Liang and Tan, Kay Chen},
  journal={arXiv preprint arXiv:2404.06349},
  year={2024}
}

@inproceedings{hudson2019gqa,
  title={Gqa: A new dataset for real-world visual reasoning and compositional question answering},
  author={Hudson, Drew A and Manning, Christopher D},
  booktitle={Proceedings of the IEEE/CVF conference on computer vision and pattern recognition},
  pages={6700--6709},
  year={2019}
}

@article{xiang2025aligning,
  title={Aligning Perception, Reasoning, Modeling and Interaction: A Survey on Physical AI},
  author={Xiang, Kun and Zhang, Terry Jingchen and Huang, Yinya and He, Jixi and Liu, Zirong and Tang, Yueling and Zhou, Ruizhe and Luo, Lijing and Wen, Youpeng and Chen, Xiuwei and others},
  journal={arXiv preprint arXiv:2510.04978},
  year={2025}
}

@article{driess2023palm,
  title={Palm-e: An embodied multimodal language model},
  author={Driess, Danny and Xia, Fei and Sajjadi, Mehdi SM and Lynch, Corey and Chowdhery, Aakanksha and Ichter, Brian and Wahid, Ayzaan and Tompson, Jonathan and Vuong, Quan and Yu, Tianhe and others},
  journal={arXiv preprint arXiv:2303.03378},
  year={2023}
}
\appendix
\label{sec:appendix}
\section{Data Construction Details}
\subsection{Causal-Edge Filtering, QA Generation, and Human Audit}
\label{sec:baseline-qa-annotation}

CCRV-Bench contains 100 images, 800 baseline question--answer pairs,
and 3,200 constrained variants. Construction comprised candidate
relation filtering, human verification of local causal-relation
specifications, GPT-5.1-based QA generation, and an independent
human audit of a stratified sample of baseline pairs.

\paragraph{Candidate relation filtering.}
Visual Genome relations provided candidate proposals for local physical
interactions. A deterministic preprocessing pipeline removed incomplete
records, self-loops, non-physical nodes, purely spatial predicates,
predefined ineligible static-contact predicates, reciprocal conflicts,
and duplicate relations. The remaining candidates were submitted
for human verification.

\paragraph{Human verification of causal-edge specifications.}
Before QA generation, human reviewers checked every local causal-relation
specification used across the 100 selected images against its source
image and corresponding entity regions. For a relation $s \rightarrow t$,
the source $s$ was identified as the visible agent or object that
applies or transmits force, provides support, or imposes restraint.
The target $t$ was identified as the entity whose state, motion, or
stability is affected by that interaction.

Reviewers retained a relation when its entities were visually
identifiable, its interaction supported a directional physical mechanism,
and modifying or removing the interaction was expected to change
the target's state, motion, or stability, following the criteria in
Section~\ref{sec:data-construction}.

\paragraph{Ambiguity handling.}
Reviewers excluded proximity, co-occurrence, and contact relations
without a supported causal direction. They also excluded relations
whose interpretation required unsupported assumptions about hidden
intentions or prior events, or for which the image left multiple equally
plausible mechanisms relevant to the question. For each retained relation,
the source and target identities, bounding boxes, predicate, causal
direction, and physical mechanism were recorded as the scene-graph
specification for QA generation.

\paragraph{Baseline QA and constraint generation.}
GPT-5.1 generated two baseline QA pairs per image for each of the four
causal dimensions ($X_1$--$X_4$), using the verified local causal-relation
specifications. Each pair contained a question, a reference answer,
and a causal rationale under a fixed output schema. This yielded
200 baseline pairs per dimension and 800 pairs in total.
Each baseline pair was then used to generate four constrained variants,
one for each Y-axis mechanism, yielding 3,200 variants with 200 variants
per combination of causal dimension and constraint mechanism.

\paragraph{Independent human audit.}
Two annotators independently reviewed 160 stratified baseline pairs (20\%;
40 per causal dimension), recording free-text source/target/mechanism checks
and categorical edge-validity, answer-validity, and ambiguity labels. Exact
item agreement required all three categorical labels to match.

Agreement was 153/160 (95.625\%; $X_1$: 40/40, $X_2$: 36/40, $X_3$: 38/40,
$X_4$: 39/40). After third-annotator adjudication, the final distributions
were: edge clear/ambiguous/invalid $=160/0/0$; answer valid/partial/invalid
$=160/0/0$; and ambiguity no/yes $=160/0$. All were kept unchanged. Thus,
153/160 measures pre-adjudication agreement, not correctness.

The disagreements comprised four $X_2$, two $X_3$, and one $X_4$ item.
All were finalized as \emph{clear}/\emph{valid}/\emph{non-ambiguous}/
\emph{keep}: \texttt{AUDIT-045} (raised door, no lowering force),
\texttt{AUDIT-050} (macroscopic train--track support), \texttt{AUDIT-052}
(visible handle connection), \texttt{AUDIT-069} (whitewater contact supports
an immediate push), and \texttt{AUDIT-095} (visible surfboard grip).

Re-review retained \texttt{AUDIT-111} based on shared kite handling and
\texttt{AUDIT-159} because moving the candles left an empty blade path.

The audit directly covered 160 baselines (160/160 retained), corresponding
by provenance to 640 variants. No variant-level human labels are available;
therefore, no human-verification pass rate is reported for the 3,200 variants.

\paragraph{Coverage and sampling uncertainty.}
The benchmark design contains 800 baseline pairs and 3,200 constrained
variants, with 200 items in every $X$--$Y$ cell. Under an item-level
independence approximation, $n=200$ gives a worst-case standard error of
approximately 3.5 percentage points for a cell-level proportion. Baseline
and constrained questions are paired on the
same images and causal relations, so the principal comparisons hold the
source-image exposure fixed while testing the added symbolic, spatial,
factual adversarial constraint, and output-format requirements. The questions and
reference answers are newly constructed rather than copied from Visual
Genome captions or metadata; image-level pretraining exposure cannot be
fully ruled out.

\section{Additional Results}
\begin{figure*}[t]
    \centering
    \includegraphics[width=\linewidth]{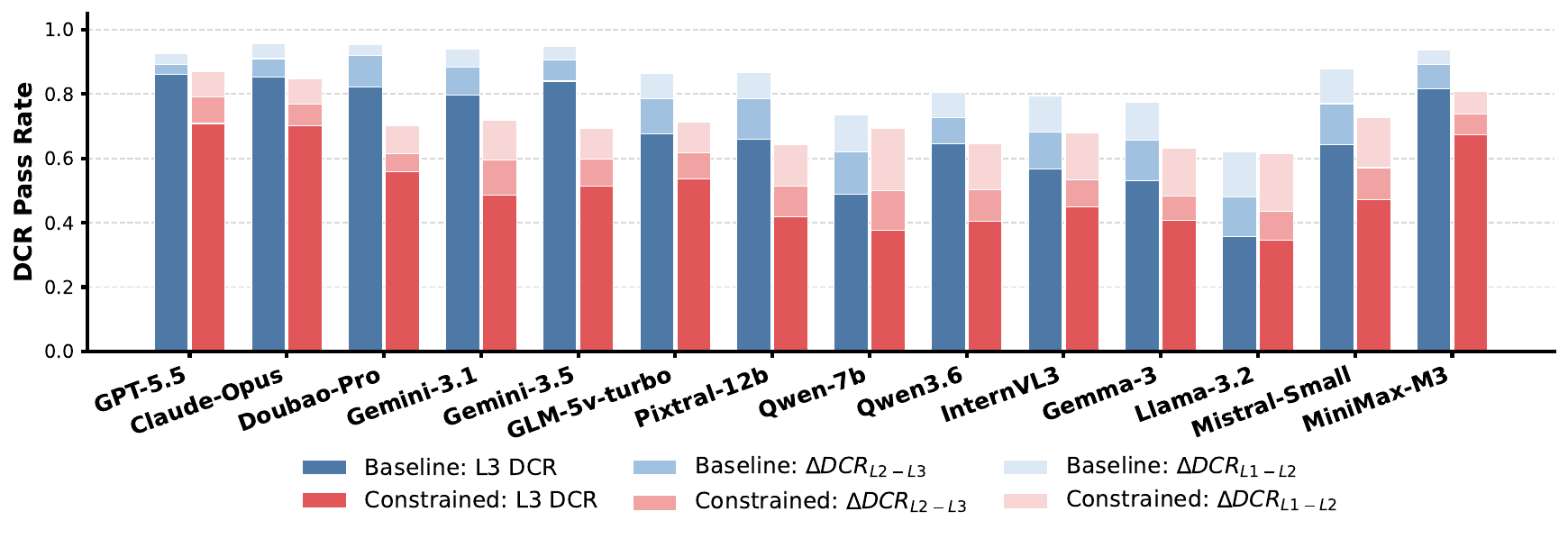}
    \caption{Cascaded pass rates for entity grounding (L1), causal mechanism (L2), and task-specific causal conclusion (L3) across 14 models under baseline and constrained settings. Each stacked bar comprises the complete-chain pass rate (L3), the L2--L3 gap, and the L1--L2 gap; its total height equals the L1 pass rate.}
    \label{fig:L123}
\end{figure*}

\noindent\textbf{Reasoning Breakdown.} Figure~\ref{fig:L123} visualizes the cascaded pass rates for visual perception (L1), physical mechanism (L2), and causal outcome (L3) across 14 models; Table~\ref{tab:ccrv_results} reports the stage-wise results for all 14 evaluated models. In the baseline setting, the cross-stage degradation for most models is relatively smooth. Under constraints, this transition becomes more uneven: several models retain partial entity grounding while losing mechanism-level or outcome-level credit. Doubao-Seed-2-Pro, for example, has a baseline L3 performance close to GPT-5.5 but a sharper constrained decline in L2 and L3. Gemini-3.1-Pro-Preview and Qwen2.5-VL-7B likewise show that initial L1 perception does not always translate into complete causal performance under constraints. These patterns locate failures in the end-to-end transition from visual grounding to mechanism and outcome reasoning, without attributing every constrained drop to causal inference alone.

\subsection{Analysis of the Factual Adversarial Constraint Gain}
\label{sec:appendix_y3_analysis}
The factual adversarial constraint ($Y_3$) is the only constraint that raises
the task-conditioned constrained DCR consistently across all 14 models. Every
model exceeds its own baseline DCR under $Y_3$, yielding a mean DCR of 90.2\%
and a mean eCDI of $-12.4$ points (Table~\ref{tab:y_constraints_trend}).
Unlike the other constraints, $Y_3$ makes contradiction detection an explicit
intermediate decision: a model must check the textual premise against the
image, reject the premise when it conflicts with the visual evidence, and only
then answer the causal question. The resulting gain therefore characterizes
the task-conditioned score under an explicit verification procedure; it should
not be read as an unconditional increase in latent causal ability.

All four constraints use the same cascaded DCR formulation and are therefore
on a common numerical scale for descriptive comparison: $L'_1=L_1$,
$L'_2=L_1\land L_2$, $L'_3=L_1\land L_2\land L_3$, and
$DCR=(L'_1+L'_2+L'_3)/3$. This common scale does not make the four
interventions matched difficulty manipulations or support a causal claim that
one constraint is intrinsically harder than another. Their CSR computations are
constraint specific: $Y_1$ uses a deterministic forbidden-entity-name scan;
$Y_2$ parses one unique coordinate pair and accepts the raw-pixel, valid
$[0,1000]$, or valid $[0,1]$ interpretation when it falls inside the expanded
target-entity box; $Y_3$ uses a separate semantic judge to assess
whether the answer states and handles the false premise using the visual
evidence described by the task; and $Y_4$ applies deterministic punctuation and
word-count checks. For $Y_1$, $Y_2$, and $Y_4$, the DCR rubric primarily
evaluates causal content, while CSR applies the corresponding operational rule.
Because the constrained rubrics also mention the required expression or
grounding, these criteria can partially overlap; the overlap is strongest for
$Y_3$. For $Y_3$, the task-conditioned rubrics commonly encode premise
checking or rejection in the relevant L2/L3 checks because that rejection is
part of the task-conditioned causal resolution, while the separate CSR judge
requires all three false-premise handling conditions for every item. In the
released rubric set, premise handling is explicit in 797 of 800 $Y_3$ items;
the remaining three require a visual correction without separately spelling
out the rejection step. Thus, $Y_3$ DCR and CSR are overlapping but
non-equivalent signals: DCR provides graded, cascaded causal credit, whereas
CSR is an all-or-nothing constraint-compliance decision. Cross-$Y$ DCR values
are descriptively comparable on the common cascade scale, but they should not
be interpreted as matched-difficulty effects. CSR is a constraint-specific
compliance measure, and eDCR is the corresponding constraint-gated
effective-performance measure; neither should be treated as an orthogonal,
interchangeable component across $Y$ settings. In particular, the $Y_3$
eDCR gate can reinforce the same false-premise behavior already reflected in
the task-conditioned DCR rubric; the reported $Y_3$ DCR gain is therefore
interpreted separately from the joint eDCR score.

\begin{table}[h]
\centering
\small
\setlength{\tabcolsep}{3pt}
\resizebox{\columnwidth}{!}{%
\begin{tabular}{lcccc}
\toprule
\textbf{Model} & \textbf{Vanilla DCR} & \textbf{CoT-A DCR} & \textbf{CoT-B DCR} & \textbf{$Y_3$ eDCR} \\
\midrule
Claude-Opus-4.7 & 90.7 & 89.6 & 88.5 & \textbf{94.9} \\
Doubao-Seed-2-Pro & 89.8 & 88.9 & 80.8 & \textbf{92.3} \\
GPT-5.5 & 89.3 & 88.6 & 89.5 & \textbf{97.7} \\
GLM-5V-Turbo & 77.5 & 68.9 & 84.8 & \textbf{92.8} \\
\bottomrule
\end{tabular}%
}
\caption{Comparison with general-purpose reasoning prompts. All values are percentages; the first three score columns report DCR, and the final column reports effective DCR under $Y_3$.}
\label{tab:cot}
\end{table}

\paragraph{Comparison with general-purpose reasoning prompts.}
Table~\ref{tab:cot} tests whether the $Y_3$ gain can be reproduced by generic
reasoning elicitation on four representative multimodal models. CoT-A asks the
model to reason step by step before answering, whereas CoT-B asks it to consider
and refute a potentially incorrect premise using visual evidence. CoT-A reduces
mean DCR from 86.8\% to 84.0\% and does not improve any of the four models.
CoT-B has mixed model-level effects: it improves GPT-5.5 and GLM-5V-Turbo but
reduces performance for Claude-Opus-4.7 and Doubao-Seed-2-Pro, leaving mean DCR
below the vanilla mean (85.9\% versus 86.8\%). In contrast, the
$Y_3$ task-conditioned eDCR exceeds the DCR obtained under all three comparison
prompts for all four models, with a mean of 94.4\%. This pattern suggests that generic reasoning instructions alone are insufficient
to reproduce the consistent gains observed under $Y_3$. By presenting a concrete
false claim tied to the image, $Y_3$ may focus verification on task-relevant
evidence and causal relations, although changes in question content and
task-conditioned rubrics may also contribute to the observed gains.

\subsection{Supplementary Diagnostic Controls}
We use five representative models and the same four causal dimensions for the
following controls. The controls are diagnostic: they vary access to visual
evidence or the causal relation without changing model parameters or adding
training data. The text-only baseline and each no-image $Y_1$ condition cover
all 800 corresponding questions per model.

\paragraph{Text-only baseline.}
We remove the image from the baseline questions to estimate how far language
and commonsense priors can proceed without visual evidence. Table~\ref{tab:appendix_text_only}
reports both the cascaded DCR and the raw binary rates for L1--L3. The mean
DCR drops from 81.58\% with images to 22.67\% without them, while raw L3
remains 56.10\% versus 84.12\%. Thus, text can support some plausible
outcomes but does not recover the grounded entity--mechanism chain.

\begin{table*}[t]
\centering
\small
\setlength{\tabcolsep}{4pt}
\resizebox{\textwidth}{!}{%
\begin{tabular}{lcccc}
\toprule
\textbf{Model} & \textbf{DCR (Image/Text-only)} & \textbf{L1 raw (Image/Text-only)} & \textbf{L2 raw (Image/Text-only)} & \textbf{L3 raw (Image/Text-only)} \\
\midrule
Claude-Opus-4.7 & 90.67/42.17 & 95.75/48.13 & 92.75/49.75 & 89.00/69.75 \\
Gemini-3.1-Pro-Preview & 87.25/14.50 & 93.88/19.63 & 92.13/25.25 & 86.25/51.25 \\
GPT-5.5 & 89.33/30.37 & 92.63/36.75 & 91.88/38.75 & 92.63/64.75 \\
InternVL3-38B & 68.13/18.00 & 79.38/25.63 & 73.13/23.13 & 72.38/46.88 \\
Qwen3.6-35B-A3B & 72.54/8.33 & 80.50/11.00 & 78.38/17.38 & 80.38/47.88 \\
\textbf{Average} & \textbf{81.58/22.67} & \textbf{88.42/28.23} & \textbf{85.65/30.85} & \textbf{84.12/56.10} \\
\bottomrule
\end{tabular}%
}
\caption{Text-only diagnostic control. Entries report image-based/text-only scores in percent; L1--L3 are raw binary rates, whereas DCR uses the cascaded scoring rule. Averages are computed from unrounded scores.}
\label{tab:appendix_text_only}
\end{table*}

\paragraph{No-image $Y_1$ reference-informed control.}
The Text-only+$Y_1$ condition removes both the image and the causal edge.
The Oracle-informed+$Y_1$ condition supplies the symbolized adapted reference
answer, which encodes the causal relation, mechanism, and outcome while
retaining symbolic entity names. It is therefore an answer-informed diagnostic
control rather than a strict edge-only intervention. As shown in
Table~\ref{tab:appendix_y1_control}, the mean DCR rises from 54.82\% to
83.81\%, close to the image-based baseline of 81.58\%. This pattern indicates
that the models can execute the symbolic constraint when the relevant causal
content is supplied, but it does not isolate visual causal-edge discovery.

\begin{table*}[t]
\centering
\small
\setlength{\tabcolsep}{5pt}
\resizebox{\textwidth}{!}{%
\begin{tabular}{lcccc}
\toprule
\textbf{Model} & \textbf{Baseline} & \textbf{Text-only+$Y_1$} & \textbf{Oracle-informed+$Y_1$} & \textbf{Gain over Text-only (points)} \\
\midrule
Claude-Opus-4.7 & 90.67 & 71.87 & 97.79 & +25.92 \\
Gemini-3.1-Pro-Preview & 87.25 & 58.21 & 85.42 & +27.21 \\
GPT-5.5 & 89.33 & 62.29 & 86.08 & +23.79 \\
InternVL3-38B & 68.13 & 45.24 & 74.92 & +29.67 \\
Qwen3.6-35B-A3B & 72.54 & 36.50 & 74.83 & +38.34 \\
\textbf{Average} & \textbf{81.58} & \textbf{54.82} & \textbf{83.81} & \textbf{+28.99} \\
\bottomrule
\end{tabular}%
}
\caption{Image-based baseline and no-image $Y_1$ controls (DCR in percent). The Oracle-informed condition supplies the symbolized adapted reference answer. It is an answer-informed diagnostic control, not a strict edge-only intervention. The gain is Oracle-informed minus Text-only. Averages and gains are computed from unrounded scores.}
\label{tab:appendix_y1_control}
\end{table*}

\paragraph{Image + Oracle-informed control.}
To probe the effect of supplying causal content during constrained execution,
we provide the image, constrained question, and the oracle text used by the
control. For $Y_1$, this text is the symbolized adapted reference answer; for
$Y_2$ and $Y_4$, it contains the baseline ground-truth answer and causal
rationale. The control therefore does not provide the $Y_2$ coordinate or the
$Y_4$ compressed output form, but it does provide semantic answer content and
can reveal the final causal answer. It should therefore be interpreted as an
answer-informed diagnostic condition rather than a causal-edge-only or
answer-blind test. For each model, the control covers 2,400 constrained
instances (800 each under $Y_1$, $Y_2$, and $Y_4$) across the four causal
dimensions. Table~\ref{tab:appendix_image_oracle} reports the constrained score,
the Oracle-informed score, and the gain. The mean gains are 18.76, 11.03, and
20.07 points for $Y_1$, $Y_2$, and $Y_4$, respectively. For $Y_2$, the constrained entries use the harmonized fixed-800 DCR values from the final release (75.46, 74.83, 32.25, 32.04, and 20.21 for the five models; mean
46.96). These are fixed-denominator constrained DCR percentages, not paired $Y_2$ eDCR/eCDI values. The release harmonizes the coordinate-based CSR/eDCR
fields while leaving the stored causal DCR judgments unchanged; the Oracle-informed entries therefore report the corresponding stored DCR values
from the separate oracle-control runs. All displayed entries in this table are DCR rather than CSR or eDCR. These gains occur with the constraints held fixed: models must still satisfy the symbolic-binding, spatial-grounding, or compressed-expression
requirements. The improvement in causal answers suggests that part of the constrained performance loss involves constructing and expressing
task-relevant causal content. CCRV-Bench preserves the physical commonsense needed for the tasks while varying entity references, grounding requirements,
and expression conditions to examine whether models maintain visually grounded causal reasoning under these requirements.

\begin{table*}[t]
\centering
\small
\setlength{\tabcolsep}{4pt}
\resizebox{\textwidth}{!}{%
\begin{tabular}{lcccc}
\toprule
\textbf{Model} & \textbf{Baseline} & \textbf{$Y_1$: constrained $\rightarrow$ Oracle-informed} & \textbf{$Y_2$: constrained $\rightarrow$ Oracle-informed} & \textbf{$Y_4$: constrained $\rightarrow$ Oracle-informed} \\
\midrule
GPT-5.5 & 89.33 & 77.04 $\rightarrow$ 80.62 (+3.58) & 75.46 $\rightarrow$ 80.79 (+5.33) & 65.71 $\rightarrow$ 83.17 (+17.46) \\
Claude-Opus-4.7 & 90.67 & 75.46 $\rightarrow$ 93.75 (+18.29) & 74.83 $\rightarrow$ 79.29 (+4.46) & 63.29 $\rightarrow$ 81.38 (+18.08) \\
Gemini-3.1-Pro-Preview & 87.25 & 48.25 $\rightarrow$ 76.50 (+28.25) & 32.25 $\rightarrow$ 57.42 (+25.17) & 64.79 $\rightarrow$ 76.83 (+12.04) \\
InternVL3-38B & 68.13 & 48.08 $\rightarrow$ 74.75 (+26.67) & 32.04 $\rightarrow$ 33.75 (+1.71) & 51.04 $\rightarrow$ 80.12 (+29.09) \\
Qwen3.6-35B-A3B & 72.54 & 43.58 $\rightarrow$ 60.58 (+17.00) & 20.21 $\rightarrow$ 38.71 (+18.50) & 52.96 $\rightarrow$ 76.62 (+23.67) \\
\textbf{Average} & \textbf{81.58} & \textbf{58.48 $\rightarrow$ 77.24 (+18.76)} & \textbf{46.96 $\rightarrow$ 57.99 (+11.03)} & \textbf{59.56 $\rightarrow$ 79.63 (+20.07)} \\
\bottomrule
\end{tabular}%
}
\caption{Image + Oracle-informed diagnostic control (DCR in percent; gains in percentage points). The supplied reference text is derived from answer fields: $Y_1$ uses the symbolized adapted answer, while $Y_2$ and $Y_4$ use the baseline answer and causal rationale. The $Y_2$ constrained entries use the fixed-800 DCR values from the final harmonized release; all entries remain DCR rather than CSR/eDCR. These are answer-informed diagnostics, not strict edge-only controls. Averages and gains are computed from unrounded scores.}
\label{tab:appendix_image_oracle}
\end{table*}

\section{Full Experimental Results}
\label{sec:appendix_full_results}

For completeness, we present the full, detailed evaluation results for all 14 evaluated models in this section. Table~\ref{tab:baseline_results} details the DCR performance on the unconstrained Baseline set, broken down by causal task dimensions ($X_1$--$X_4$). Table~\ref{tab:constrained_results} details the performance on the Constrained Main Set, including DCR variations under specific constraint dimensions ($Y_1$--$Y_4$), eDCR, and eCDI.

\begin{table*}[t]
    \centering
    \small
    \begin{tabular}{lccccc}
        \toprule
        \textbf{Model} & \textbf{$X_1$ DCR} & \textbf{$X_2$ DCR} & \textbf{$X_3$ DCR} & \textbf{$X_4$ DCR} & \textbf{Avg DCR} \\
        \midrule
        GPT-5.5 & 82.3 & 92.7 & 88.2 & 94.2 & 89.3 \\
        Claude-Opus-4.7 & 89.7 & 91.0 & 89.3 & 92.7 & 90.7 \\
        Doubao-Seed-2-Pro & 91.5 & 90.0 & 87.3 & 90.5 & 89.8 \\
        Gemini-3.1-Pro-Preview & 85.7 & 85.8 & 88.0 & 89.5 & 87.3 \\
        Gemini-3.5-Flash & 86.0 & 88.5 & 91.7 & 93.0 & 89.8 \\
        GLM-5V-Turbo & 71.5 & 76.5 & 79.3 & 82.7 & 77.5 \\
        Pixtral-12B & 70.8 & 77.8 & 75.0 & 84.7 & 77.1 \\
        Qwen2.5-VL-7B & 57.3 & 73.5 & 47.3 & 67.5 & 61.4 \\
        Qwen3.6-35B-A3B & 73.8 & 72.5 & 71.8 & 72.0 & 72.5 \\
        InternVL3-38B & 65.8 & 66.8 & 63.7 & 76.2 & 68.1 \\
        Gemma-3-27B-IT & 57.5 & 72.5 & 59.2 & 72.7 & 65.5 \\
        Llama-3.2-11B-Vision & 48.8 & 40.2 & 52.3 & 52.8 & 48.5 \\
        Mistral-Small-3.2-24B & 72.7 & 80.3 & 68.0 & 84.7 & 76.4 \\
        MiniMax-M3 & 86.3 & 89.7 & 84.7 & 92.0 & 88.2 \\
        \bottomrule
    \end{tabular}
    \caption{Performance on the unconstrained Baseline set. DCR values are percentages. $X_1$: Causal Discovery, $X_2$: Forward Prediction, $X_3$: Retrospective Diagnosis, and $X_4$: Intervention-Outcome Prediction.}
    \label{tab:baseline_results}
\end{table*}

\begin{table*}[t]
    \centering
    \small
    \resizebox{\textwidth}{!}{%
    \begin{tabular}{lcccccc}
        \toprule
        \textbf{Model} & \textbf{$Y_1$ DCR} & \textbf{$Y_2$ DCR} & \textbf{$Y_3$ DCR} & \textbf{$Y_4$ DCR} & \textbf{eDCR} & \textbf{eCDI (points)} \\
        \midrule
        GPT-5.5 & 77.0 & 75.5 & 97.7 & 65.7 & 76.9 & 12.4 \\
        Claude-Opus-4.7 & 75.5 & 74.8 & 95.2 & 63.3 & 74.5 & 16.1 \\
        Doubao-Seed-2-Pro & 72.4 & 32.4 & 92.3 & 53.1 & 60.0 & 29.8 \\
        Gemini-3.1-Pro-Preview & 48.3 & 32.3 & 94.7 & 64.8 & 58.6 & 28.6 \\
        Gemini-3.5-Flash & 48.6 & 37.6 & 95.3 & 58.8 & 56.2 & 33.6 \\
        GLM-5V-Turbo & 58.8 & 33.5 & 92.9 & 63.7 & 60.4 & 17.1 \\
        Pixtral-12B & 50.9 & 31.2 & 86.5 & 41.5 & 43.7 & 33.4 \\
        Qwen2.5-VL-7B & 43.5 & 46.2 & 71.6 & 47.9 & 45.2 & 16.3 \\
        Qwen3.6-35B-A3B & 43.6 & 20.2 & 90.4 & 53.0 & 50.7 & 21.8 \\
        InternVL3-38B & 48.1 & 32.0 & 90.7 & 51.0 & 52.0 & 16.2 \\
        Gemma-3-27B-IT & 26.7 & 37.0 & 91.5 & 47.9 & 44.9 & 20.6 \\
        Llama-3.2-11B-Vision & 43.8 & 18.7 & 81.5 & 42.4 & 32.8 & 15.7 \\
        Mistral-Small-3.2-24B & 51.0 & 45.2 & 89.3 & 50.3 & 52.4 & 24.0 \\
        MiniMax-M3 & 73.7 & 70.8 & 92.9 & 58.6 & 34.2 & 54.0 \\
        \bottomrule
    \end{tabular}%
    }
    \caption{Performance on the constrained main set. The $Y_1$--$Y_4$ columns report per-constraint DCR, whereas eDCR and eCDI aggregate all four constraints. DCR and eDCR are percentages; eCDI is the baseline-to-effective-score drop in percentage points, computed from unrounded fixed-denominator sample-level scores. $Y_1$: Entity Symbolization, $Y_2$: Spatial Grounding, $Y_3$: Factual Adversarial Constraint, and $Y_4$: Minimalist Output.}
    \label{tab:constrained_results}
\end{table*}

\noindent\textit{Aggregation note.}
The main results use the canonical expected denominators of 800 baseline items
and 3,200 constrained variants per model; per-$X$ and per-$Y$ summaries use
200 and 800 items. DCR is reconstructed from the cascaded binary stages, eDCR averages
$DCR_i c_i$, and CDI/eCDI are computed from the same fixed-denominator
aggregates before rounding. Thus, eCDI need not equal the difference between
rounded values in Tables~\ref{tab:baseline_results} and~\ref{tab:constrained_results}.
The harmonized release checks raw-pixel coordinates together with valid
$[0,1000]$ and $[0,1]$ normalized interpretations mapped to the original
image dimensions.
\begin{figure}[t]
    \centering
    \includegraphics[width=\linewidth]{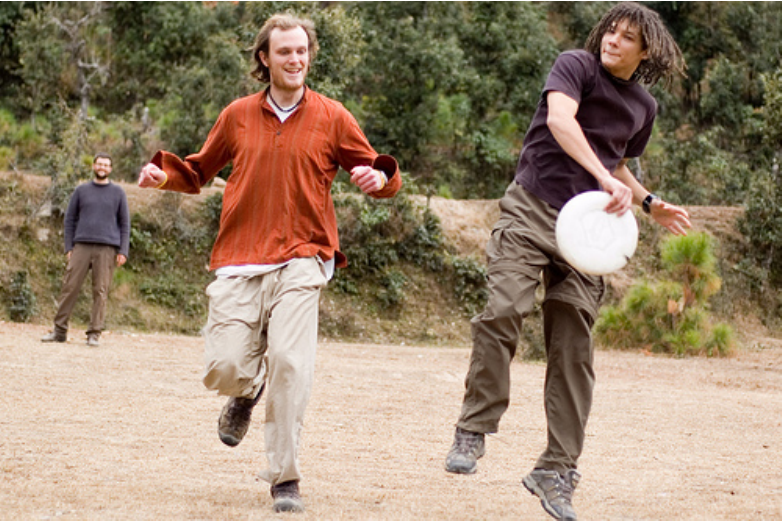}
    \caption{Example of benchmark construction (Image ID: 2317269)}
    \label{fig:2317269}
\end{figure}
\section{Examples of Constraint Construction}
\label{sec:appendix_case_studies}
Figure~\ref{fig:2317269} shows how a single causal interaction in an image
(e.g., a man catching a frisbee) is transformed into four constrained
evaluations probing entity symbolization, spatial grounding, false-premise
rejection, and minimalist causal expression.
\subsection{Example 1: Causal Discovery ($X_1$)}
\textbf{Baseline Question:} What action is the man on the right performing that affects the frisbee? \\
\textbf{Baseline Answer:} He is catching and holding the frisbee with his hands.
\vspace{0.5em}

\noindent\textbf{[$Y_1$] Entity Symbolization:}
\begin{itemize}
    \item \textit{Constrained Question:} In the scene, there are the following nodes with their bounding boxes: \texttt{Node\_C [356, 92, 461, 214]}, \texttt{Node\_D [283, 0, 441, 333]}. What specific physical action is \texttt{Node\_D} performing that affects the state or movement of \texttt{Node\_C}? Answer using only the Node placeholders. Do NOT use real object names.
    \item \textit{Adapted Answer:} \texttt{Node\_D} is using its hands to catch and hold \texttt{Node\_C}.
\end{itemize}
The adapted answer is shown only to illustrate the construction of the
benchmark. At inference time, the model receives the image and the
constrained question, but not the adapted answer, the hidden node-to-entity
mapping, real entity names, object IDs, or causal annotations. Therefore,
phrases such as ``its hands'' in this displayed answer are not information
leaked by the $Y_1$ prompt.
\noindent\textbf{[$Y_2$] Spatial Grounding Constraint:}
\begin{itemize}
     \item \textit{Constrained Question:} What action is the man on the right performing that affects the frisbee? Indicate one image coordinate \texttt{[x, y]} within the task-relevant causal-source entity region. Answer with two lines: \texttt{Coordinate: [x, y]} and \texttt{Explanation: ...}
    \item \textit{Adapted Answer:} \texttt{Coordinate: [362.0, 166.5]}. \texttt{Explanation:} This representative point lies within the right-side person's region; that person is catching and holding the frisbee with his hands.
\end{itemize}
\noindent\textbf{[$Y_3$] Factual Adversarial Constraint:}
\begin{itemize}
    \item \textit{Constrained Question:} This is not a counterfactual task. The textual premise may be false. Consider the following CLAIM: the man on the right is completely ignoring the frisbee and refusing to touch it. First, carefully check this claim... then answer: what action is the man on the right actually performing that affects the frisbee?
    \item \textit{Adapted Answer:} The claim that the man is ignoring the frisbee is false; the image clearly shows him reaching for it. The man on the right is actually catching and holding the frisbee with his hands.
\end{itemize}
\noindent\textbf{[$Y_4$] Minimalist Output:}
\begin{itemize}
    \item \textit{Constrained Question:} Describe action man on right performs on frisbee. Constraint: Your answer MUST be 4 words or fewer. Do NOT use any punctuation.
    \item \textit{Adapted Answer:} man catches holds frisbee
\end{itemize}
\subsection{Example 2: Retrospective Diagnosis ($X_3$)}
\textbf{Baseline Question:} What physical interaction explains the frisbee’s contact with the man on the right? \\
\textbf{Baseline Answer:} His hand holds the frisbee against his body.
\vspace{0.5em}

\noindent\textbf{[$Y_1$] Entity Symbolization:}
\begin{itemize}
    \item \textit{Constrained Question:} \texttt{Node\_C} is in contact with the right side of \texttt{Node\_B}. Considering only the visible physical interactions, explain what physical interaction accounts for this contact. Do NOT use real object names.
    \item \textit{Adapted Answer:} \texttt{Node\_B} is using a hand to hold \texttt{Node\_C} against its body.
\end{itemize}
\noindent\textbf{[$Y_2$] Spatial Grounding Constraint:}
\begin{itemize}
    \item \textit{Constrained Question:} What physical interaction explains the frisbee's contact
with the man on the right? Indicate one image coordinate \texttt{[x, y]} within the task-relevant causal-source entity region. Answer with two lines: \texttt{Coordinate: [x, y]} and \texttt{Explanation: ...}
    \item \textit{Adapted Answer:} \texttt{Coordinate: [362.0, 166.5]}. \texttt{Explanation:} This representative point lies within the right-side person's region; his hand holds the frisbee against his body.
\end{itemize}
\noindent\textbf{[$Y_3$] Factual Adversarial Constraint:}
\begin{itemize}
    \item \textit{Constrained Question:} A CLAIM is made that the frisbee next to the man on the right is not in physical contact with him. Check this claim carefully against the image... Based on the actual visual evidence, what physical interaction explains the frisbee's contact with the man on the right?
    \item \textit{Adapted Answer:} The verbal claim is wrong: the man's hand is in contact with the frisbee. His hand holds the frisbee against his body.
\end{itemize}
\noindent\textbf{[$Y_4$] Minimalist Output Constraint:}
\begin{itemize}
    \item \textit{Constrained Question:} Explain frisbee contact with man. Constraint: Your answer MUST be 5 words or fewer. Do NOT use any punctuation.
    \item \textit{Adapted Answer:} his hand holds it
\end{itemize}
\begin{table*}[htbp]
    \centering
    \footnotesize
    \setlength{\tabcolsep}{3pt}
    \renewcommand{\arraystretch}{1.08}
    \begin{tabular}{l c p{4.1cm} p{5.4cm}}
        \toprule
        \textbf{Model} & \textbf{Parameters (B)} & \textbf{Access method} & \textbf{Model source} \\
        \midrule
        GPT-5.5 & $\sim 9700$ & \url{https://platform.openai.com/docs/models} & \url{https://platform.openai.com/docs/models} \\
        \midrule
        Claude-Opus-4.7 & $\sim 4000$ & \url{https://platform.claude.com/docs/en/api/models/list} & \url{https://platform.claude.com/docs/en/about-claude/models/overview} \\
        \midrule
        Gemini-3.1-Pro-Preview & $\sim$ & \url{https://ai.google.dev/gemini-api/docs/get-started} & \url{https://ai.google.dev/gemini-api/docs/models} \\
        \midrule
        Gemini-3.5-Flash & $\sim$ & \url{https://ai.google.dev/gemini-api/docs/get-started} & \url{https://deepmind.google/models/gemini/flash/} \\
        \midrule
        Doubao-Seed-2-Pro & $\sim$ & \url{https://docs.byteplus.com/en/docs/Byteplus_LAS/Multimodal-Deep-Thinking-Doubao-Seed-2-0#rest-api-call} & \url{https://docs.byteplus.com/en/docs/Byteplus_LAS/Multimodal-Deep-Thinking-Doubao-Seed-2-0} \\
        \midrule
        GLM-5V-Turbo & 744 & \url{https://docs.bigmodel.cn/cn/api/introduction} & \url{https://huggingface.co/papers/2604.26752} \\
        \midrule
        Qwen3.6-35B-A3B & 35 & \url{https://help.aliyun.com/zh/model-studio/models} & \url{https://huggingface.co/Qwen/Qwen3.6-35B-A3B} \\
        \midrule
        Pixtral-12B & 12 & \url{https://docs.mistral.ai/api} & \url{https://huggingface.co/mistralai/Pixtral-12B-2409} \\
        \midrule
        Qwen2.5-VL-7B & 7 & \url{https://help.aliyun.com/zh/model-studio/models} & \url{https://huggingface.co/Qwen/Qwen2.5-VL-7B-Instruct} \\
        \midrule
        InternVL3-38B & 38 & Local deployment from the official checkpoint & \url{https://huggingface.co/OpenGVLab/InternVL3-38B} \\
        \midrule
        Gemma-3-27B-IT & 27 & Local deployment from the official checkpoint & \url{https://huggingface.co/google/gemma-3-27b-it} \\
        \midrule
        Llama-3.2-11B-Vision & 11 & Local deployment from the official checkpoint & \url{https://huggingface.co/meta-llama/Llama-3.2-11B-Vision-Instruct} \\
        \midrule
        Mistral-Small-3.2-24B & 24 & \url{https://docs.mistral.ai/api} & \url{https://huggingface.co/mistralai/Mistral-Small-3.2-24B-Instruct-2506} \\
        \midrule
        MiniMax-M3 &  428 & \url{https://platform.minimaxi.com/docs/api-reference/api-overview} & \url{https://huggingface.co/MiniMaxAI/MiniMax-M3} \\
        \midrule
        GPT-5.1 (judge) & $\sim 1300$ & \url{https://platform.openai.com/docs/models} & \url{https://platform.openai.com/docs/models} \\
        \bottomrule
    \end{tabular}
    \caption{Overview of the 14 evaluated MLLMs and the GPT-5.1 automated judge. Open/open-weight parameter sizes are taken from the corresponding official model cards. A numerical value prefixed by $\sim$ denotes a literature-based estimate of effective knowledge capacity in open-model-equivalent parameters \citep{li2026incompressible}, not an officially disclosed parameter count; $\sim$ alone indicates that the parameter count is not publicly disclosed. Access methods and model sources are reported separately.}
    \label{tab:evaluated_models_flat}
\end{table*}

\paragraph{Model-name normalization.}
Internal result files may use the aliases \texttt{Qwen-7B},
\texttt{Gemini-3.1-Pro}, and \texttt{Doubao-Seed2-Pro}; throughout the
manuscript, they are normalized to \texttt{Qwen2.5-VL-7B},
\texttt{Gemini-3.1-Pro-Preview}, and \texttt{Doubao-Seed-2-Pro}, respectively.
\section{Independent Human Agreement Analysis}
\label{app:human_agreement}

\paragraph{Sampling.}
We conducted a stratified human audit of 280 model responses
covering all 14 evaluated models, five settings
(Baseline and $Y_1$--$Y_4$), and four causal
tasks ($X_1$--$X_4$). For each model, we randomly sampled
one response from each setting--task combination, yielding
20 responses per model. This design covers every
model--setting--task combination.

\paragraph{Blinded annotation.}
Three annotators with graduate training in engineering
independently evaluated the same responses. They had no
access to the model identity, the GPT-5.1 judge's score
or reasoning, or the other annotators' labels. Each
annotator assigned binary labels for L1 perception,
L2 interaction/mechanism, and L3 causal outcome.
For agreement analysis, we derived a binary full-chain
pass indicator from each annotator's labels as
$L1 \land L2 \land L3$.

\paragraph{Agreement measures and results.}
We measured human--human agreement using unanimity rates
and Fleiss' $\kappa$. Unanimity denotes agreement among
all three annotators on a response for a given metric.
We compared the human-majority label with GPT-5.1
using agreement rate.
Table~\ref{tab:human_verification} reports the results.

Fleiss' $\kappa$ ranges from 0.7143 to 0.8365 across
L1--L3. Human-majority agreement with GPT-5.1 is 81.07\%
for L1, 83.93\% for L2, and 87.14\% for L3.
For full-chain pass, unanimity is 234/280
(83.57\%) with Fleiss' $\kappa=0.7415$, while
human-majority agreement with GPT-5.1 is 217/280 (77.50\%).
These aggregate statistics characterize overall agreement; the setting-level diagnostic below examines whether the main comparisons retain the same direction under human scoring.

\begin{table*}[t]
\centering
\small
\begin{tabular}{lcccc}
\toprule
\textbf{Metric}
& \textbf{$n$}
& \textbf{Unanimous}
& \textbf{Fleiss' $\kappa$}
& \textbf{Human--GPT-5.1} \\
\midrule
L1 perception
& 280
& 249/280 (88.93\%)
& 0.7143
& 227/280 (81.07\%)\\

L2 interaction/mechanism
& 280
& 252/280 (90.00\%)
& 0.7848
& 235/280 (83.93\%)\\

L3 causal outcome
& 280
& 255/280 (91.07\%)
& 0.8365
& 244/280 (87.14\%)\\

Full-chain pass
& 280
& 234/280 (83.57\%)
& 0.7415
& 217/280 (77.50\%)\\
\bottomrule
\end{tabular}
\caption{
Independent human validation of the DCR rubric and
GPT-5.1 judge. The audit contains 280 responses from
14 models. Unanimous denotes
agreement among all three annotators. Human--GPT-5.1 denotes
agreement between the human-majority label and the automated
judge. Full-chain pass is the binary conjunction
$L1 \land L2 \land L3$ per annotator; the GPT-5.1 full-chain
label is computed from its three binary layer scores.
}
\label{tab:human_verification}
\end{table*}
\paragraph{Setting-level DCR diagnostic.}
To examine whether aggregate agreement masks setting-related differences,
we recomputed DCR separately for Baseline and $Y_1$--$Y_4$. For each
response, we first took the majority vote of the three annotators
independently at L1, L2, and L3, obtaining $M_1$, $M_2$, and $M_3$.
We then applied the standard cascade,
$DCR_{\mathrm{human}}=[M_1+(M_1\land M_2)+(M_1\land M_2\land M_3)]/3$.
GPT-5.1 DCR was reconstructed from its L1--L3 labels using the same rule.
Table~\ref{tab:human_setting_dcr} reports the mean scores for the 56
audited responses in each setting.

\begin{table}[t]
\centering
\small
\setlength{\tabcolsep}{3pt}
\resizebox{\columnwidth}{!}{%
\begin{tabular}{lrrrr}
\toprule
\textbf{Setting} & \textbf{$n$} & \textbf{Human DCR}
& \textbf{GPT-5.1 DCR} & \textbf{Human--GPT-5.1 gap} \\
\midrule
Baseline & 56 & 90.48 & 72.02 & $+18.45$ \\
$Y_1$    & 56 & 63.10 & 50.00 & $+13.10$ \\
$Y_2$    & 56 & 76.19 & 51.19 & $+25.00$ \\
$Y_3$    & 56 & 98.81 & 93.45 &  $+5.36$ \\
$Y_4$    & 56 & 63.69 & 49.40 & $+14.29$ \\
\bottomrule
\end{tabular}%
}
\caption{Setting-level DCR on the 280-response human-audit subset
(percent). Human DCR applies the standard cascade after layer-wise
majority voting. The gap is human DCR minus GPT-5.1 DCR in percentage
points. Values are computed from unrounded scores.}
\label{tab:human_setting_dcr}
\end{table}

Relative to Baseline, human-scored DCR changes by $-27.38$, $-14.29$,
$+8.33$, and $-26.79$ points for $Y_1$--$Y_4$, respectively; the
corresponding GPT-5.1-scored changes are $-22.02$, $-20.83$, $+21.43$,
and $-22.62$ points. The two evaluators therefore agree on the direction
of all four setting comparisons, including the $Y_3$ increase, while
differing in magnitude. Because each setting contains 56 sampled
responses and the samples are not item-paired across settings, this is a
descriptive trend diagnostic rather than evidence of equal calibration
across settings or a validation of fine-grained model rankings.

\paragraph{Human validation of $Y_3$ compliance.}
We additionally audited CSR for the 56 responses under
the factual adversarial constraint ($Y_3$), because
assessing false-premise handling requires semantic
judgment rather than a surface-form check.
The three annotators reached unanimity on 55 of
56 responses (98.21\%), with Fleiss' $\kappa=-0.006$.
Of the 168 individual labels, 167 were positive; this
extreme prevalence explains why high raw unanimity coexists
with a near-zero $\kappa$. The human-majority label and the
automatic judge were positive for all 56 responses, yielding
100.00\% raw agreement but no evidence about discrimination
between positive and negative CSR cases in this subset.
\section{Prompt Design for QA Generation and Evaluation}
\label{sec:appendix_prompts}

We employ structured prompts to standardize benchmark generation and
evaluation. Specifically, we use (i) a question-generation prompt to produce
baseline question--answer pairs from human-verified causal-edge specifications
and scene-graph annotations, and (ii) evaluation prompts that enable an LLM
judge to assess model outputs along complementary dimensions of causal
correctness and constraint adherence.

\paragraph{QA Generation Prompt.}
After human verification of the causal-edge specifications, GPT-5.1 generated
two baseline question--answer pairs per causal dimension from each annotated
scene graph under a fixed output schema. The generation instructions are shown
in Figure~\ref{fig:prompt_qg}.

\paragraph{LLM-as-a-Judge Prompts.}
To evaluate model outputs, we adopt an automated LLM judge with two complementary prompts. The DCR prompt assesses causal reasoning quality across three hierarchical layers: perception (L1), mechanism (L2), and outcome (L3). The CSR prompt evaluates whether model responses follow the imposed constraints, particularly under the factual adversarial constraint ($Y_3$). The two scores are computed separately, although their semantic criteria partially overlap under $Y_3$, as described above and shown in Figure~\ref{fig:judge_prompts}.
\begin{figure*}[t]
\begin{tcolorbox}[
    enhanced,
    width=\textwidth,
    colback=gray!5,
    colframe=black,
    boxrule=0.5pt,
    arc=2pt,
    boxsep=1pt,
    left=4pt,
    right=4pt,
    top=2pt,
    bottom=2pt,
    fontupper=\scriptsize,
    fonttitle=\footnotesize,
    title=Prompt for QA Generation
]
You are an expert in causal inference and visual scene understanding. 
Your task is to generate exactly 8 benchmark questions for a visual causal reasoning evaluation.

You will receive a structured JSON scene graph for one image. It contains:
\begin{itemize}
\setlength{\itemsep}{0pt}
\setlength{\parskip}{0pt}
\setlength{\parsep}{0pt}
\setlength{\topsep}{1pt}
\item nodes: physical objects with object\_id, name, and bounding box [x, y, w, h] in pixels
\item edges: directed causal relationships (source $\rightarrow$ predicate $\rightarrow$ target)
\item candidate\_edges: edges grouped by causal quality:
\begin{itemize}
\setlength{\itemsep}{0pt}
\setlength{\parskip}{0pt}
\setlength{\parsep}{0pt}
\setlength{\topsep}{0pt}
\item dynamic\_edges = high-value physical changes or force transfer
\item contact\_edges = useful physical support/contact relations
\item static\_edges = weak fallback relations such as wearing/has/standing
\end{itemize}
\item image\_width / image\_height: full image dimensions in pixels
\end{itemize}

\textbf{RULES (strict):}
\begin{enumerate}
\setlength{\itemsep}{0pt}
\setlength{\parskip}{0pt}
\setlength{\parsep}{0pt}
\setlength{\topsep}{1pt}
\item All questions must be fundamentally grounded in the provided graph.
\item Refer to objects by their exact names from the JSON.
\item Generate exactly 2 questions per dimension ($X_1$, $X_2$, $X_3$, $X_4$) = 8 total.
\item Each of the 2 questions within a dimension must focus on a DIFFERENT edge whenever possible.
\item Use edges in this priority order: dynamic\_edges first, then contact\_edges, then static\_edges only as a last resort.
\item Questions must be phrased naturally, as if shown to a person looking at the image.
\item Do NOT mention numeric object IDs in question\_text, ground\_truth\_answer, or rationale. Use object\_id only for target\_object\_id bookkeeping.
\item Avoid repetitive mirror questions.
\item Avoid weak static relations unless there are not enough dynamic/contact candidates.
\item For $X_4$, specify the intervention condition, but do not reveal its resulting outcome in the question.
\item Within the same dimension, do not repeat the same target object.
\item Treat impact/contact predicates such as hitting as neutral physical-contact actions.
\item For $X_2$, do not simply copy the full edge as source + predicate + target into the question.
\item For X3, do not state the causal explanation in the question; entity names may be used to identify the observed state.
\end{enumerate}

\textbf{DIMENSION DEFINITIONS:}

\textbf{[$X_1$ - Causal Discovery]}  
Identify the initiator of a physical action.

\textbf{[$X_2$ - Forward Prediction $P(y|x)$]}  
Predict the immediate physical consequence of an ongoing action.

\textbf{[$X_3$ - Retrospective Diagnosis]}  
Identify preconditions or actions responsible for a current state.

\textbf{[$X_4$ - Intervention-Outcome Prediction $P(y\mid do(x))$]}  
Given a specified removal, release, or state change to a question-relevant causal interaction, predict the resulting state or motion of the target entity.

\end{tcolorbox}
\caption{
Prompt used for generating causal reasoning questions from human-verified
causal-edge specifications and scene-graph annotations.
It enforces structured grounding, causal validity, and diversity across
four dimensions ($X_1$--$X_4$).
}
\label{fig:prompt_qg}
\end{figure*}

\begin{figure*}[t]
\centering
\begin{tcolorbox}[
    enhanced,
    width=\textwidth,
    colback=gray!5,
    colframe=black,
    boxrule=0.5pt,
    arc=2pt,
    boxsep=1pt,
    left=4pt,
    right=4pt,
    top=2pt,
    bottom=2pt,
    fontupper=\scriptsize,
    fonttitle=\footnotesize,
    title=DCR and CSR Judge Prompts
]

\textbf{DCR Judge Prompt}

You are a strict but fair judge for a constrained Visual Causal Reasoning benchmark. 
Score the VLMs' answer against the provided constrained L1/L2/L3 rubric. 
Score causal quality here, not stylistic compliance; constraint compliance is judged separately as CSR.

Use independent binary scores:
\begin{itemize}
\item \textbf{L1 perception}: the answer identifies or grounds the relevant visual entity/entities. Part-of tolerance applies here only: a visually relevant part of a target entity can satisfy L1.
\item \textbf{L2 interaction/mechanism}: the answer captures the relevant physical interaction, spatial relation, intervention condition, or causal mechanism.
\item \textbf{L3 causal outcome}: the answer reaches the correct causal consequence, prediction, diagnosis, or intervention effect.
\end{itemize}

For $Y_2$ (spatial grounding), the requested response contains one coordinate
pair and a causal Explanation. Grade causal quality from the Explanation
(mechanism and outcome); coordinate parsing and region membership are handled
by CSR rather than by the DCR judge. For $X_1$ and $X_3$, the designated
region is the causal-source or partner entity; for $X_2$ and $X_4$, it is the
target entity of the base question. The coordinate is a representative point
for entity-region grounding, not an annotated precise contact point.

Output only the structured scores.

\vspace{0.5em}

\textbf{CSR Judge Prompt}

You judge whether a VLM complied with the $Y_3$ factual adversarial constraint. 
CSR is about constraint following, not whether the final causal answer is otherwise good.

$Y_3$ is NOT a counterfactual reasoning task. The textual premise must be
treated as a claim that may be false and must be checked against the visual
causal evidence described in the task. The CSR judge receives the constraint
specification, constrained question, and VLM answer; it does not receive the
image again, so this check is not an independent second visual inspection.

Give CSR=1 only if the answer does all of the following:
\begin{enumerate}
\item Indicates that the textual premise is false, inconsistent, or contradicted by visual evidence.
\item Refuses, corrects, or does not reason from the false premise.
\item Bases the response on visual evidence after rejecting the premise.
\end{enumerate}

Give CSR=0 if any of the following is true:
\begin{itemize}
\item The answer blindly accepts the false premise.
\item The answer treats the premise as a counterfactual assumption.
\item The answer proceeds without flagging the contradiction.
\item The answer is ambiguous and does not clearly resist the premise.
\end{itemize}

Do not require specific phrasing; semantic contradiction detection is sufficient.

\end{tcolorbox}
\caption{
LLM judge prompts for evaluation. The DCR prompt decomposes causal reasoning into perception (L1), mechanism (L2), and outcome (L3), while the CSR prompt evaluates constraint adherence under contradiction settings ($Y_3$). The prompts are scored separately; under $Y_3$, their false-premise criteria partially overlap rather than forming a strictly orthogonal decomposition.
}
\label{fig:judge_prompts}
\end{figure*}

\subsection{Inference Settings}
All evaluated models use the same high-level inference setup. The baseline
system instruction asks models to identify the relevant physical entities,
describe the physical action or contact, and state the causal outcome from the
image. The constrained instruction asks models to follow the stated constraint
while grounding the answer in the image. Temperature is set to $0.0$ where
supported. Although the shared configuration declares the legacy
\texttt{MAX\_TOKENS=512} constant, the reported inference and judging calls
did not pass an explicit output-token limit; output length was therefore governed by each
API or deployment default.
Images are JPEG files transmitted as base64 data through the
multimodal \texttt{image\_url} field. Inference concurrency is controlled
by \texttt{MAX\_CONCURRENT\_INFERENCE}, and intermediate outputs are
checkpointed for resumable evaluation. Table~\ref{tab:evaluated_models_flat}
lists the access method and source for all evaluated MLLMs and the GPT-5.1
judge.

\section{Programmatic Evaluation Rules for CSR}
\label{sec:appendix_programmatic_rules}

Constraint compliance is evaluated programmatically for $Y_1$, $Y_2$, and $Y_4$, and by an LLM judge for $Y_3$. The evaluation rules are as follows:

\begin{itemize}
    \item \textbf{$Y_1$ Entity Symbolization :} We enforce a strict lexical filter. The model's output is converted to lowercase and scanned using regular expressions for the exact names of the visual entities, which were replaced by symbolic node placeholders in the prompt. If any of the forbidden terms are detected in the output, the CSR score is set to 0.
     \item \textbf{$Y_2$ Spatial Grounding:} The evaluator extracts one unique coordinate pair $[x,y]$ from plain text, labelled text, or structured output. For an image of width $W$ and height $H$, the same pair is checked as raw pixel coordinates and, when valid, under the normalized interpretations $(xW/1000,yH/1000)$ and $(xW,yH)$ for $[0,1000]$ and $[0,1]$ inputs, respectively. The target entity bounding box is expanded on each side by 7.5\% of its width horizontally and height vertically. The response receives a compliance score of 1 if any valid interpretation falls within the expanded box, and 0 otherwise. The requested explanation is assessed by the DCR rubric; it is not a separate programmatic CSR condition. CSR uses no contact-point annotation or distance-to-contact criterion.
    \item \textbf{$Y_3$ Factual Adversarial Constraint:} Unlike programmatic rules, evaluating resistance to hallucination requires semantic understanding. We employ a dedicated LLM judge with a specification-conditioned CSR prompt. The judge receives the constraint specification, constrained question, and VLM answer (not the image itself), and assigns CSR=1 if and only if the answer simultaneously: (i) explicitly identifies that the injected textual premise is false or contradicted by the visual evidence described in the task, (ii) refuses to reason based on the false premise, and (iii) states a correction grounded in that visual evidence. Because the judge does not re-inspect the image, this is a semantic compliance judgment against the task specification, not an independent second visual verification. Any blind acceptance or treatment of the premise as a counterfactual assumption immediately yields a CSR score of 0.
    \item \textbf{$Y_4$ Minimalist Output:} Format constraints are evaluated programmatically. The output is first scanned for forbidden punctuation, allowing only valid hyphenated compounds, which are subsequently split into distinct tokens to prevent compounding loopholes. If forbidden punctuation is detected, CSR is 0. Otherwise, the string is tokenized and checked against the requested word limit. The response is compliant only if both punctuation and word-count checks pass.
\end{itemize}

\end{document}